\documentclass{article}
\usepackage{ijcai23}

\usepackage{times}
\usepackage{soul}
\usepackage{url}
\usepackage[hidelinks]{hyperref}
\usepackage[utf8]{inputenc}
\usepackage[small]{caption}
\usepackage{graphicx}
\usepackage{amsmath}
\usepackage{amsthm}
\usepackage{booktabs}
\usepackage{algorithm}
\usepackage{algorithmic}
\usepackage{subfigure}
\usepackage{amsfonts}
\usepackage{multirow}
\usepackage{bm}
\usepackage{stfloats}
\usepackage{xr}
\usepackage{lipsum}
\usepackage[group-separator={,},group-minimum-digits=4]{siunitx}

\usepackage[switch]{lineno}

\title{Structural Imbalance Aware Graph Augmentation Learning}

\author{
	Zulong Liu$^1$
	\and
	Ke-jia Chen$^{1, }$\footnote{Corresponding author}\and
	Zheng Liu$^{1}$
	\affiliations
	$^1$School of Computer Science and Technology, Nanjing University of Posts and Telecommunications\\
	\emails
	{1021041210, chenkj, zliu}@njupt.edu.cn
}

\begin{document}
\maketitle

\begin{abstract}
    Graph machine learning (GML) has made great progress in node classification, link prediction, graph classification and so on. However, graphs in reality are often structurally imbalanced, that is, only a few hub nodes have a denser local structure and higher inﬂuence. The imbalance may compromise the robustness of existing GML models, especially in learning tail nodes. This paper proposes a selective graph augmentation method (SAug) to solve this problem. Firstly, a Pagerank-based sampling strategy is designed to identify hub nodes and tail nodes in the graph. Secondly, a selective augmentation strategy is proposed, which drops the noisy neighbors of hub nodes on one side, and discovers the latent neighbors and generates pseudo neighbors for tail nodes on the other side. It can also alleviate the structural imbalance between two types of nodes. Finally, a GNN model will be retrained on the augmented graph. Extensive experiments demonstrate that SAug can significantly improve the backbone GNNs and achieve superior performance to its competitors of graph augmentation methods and hub/tail aware methods. \footnotetext{\emph{Preprint Under Review}}
\end{abstract}

\section{Introduction}

The graph is a data structure widely used to model relational and structural data in real-world networks, such as social networks, protein networks, semantic networks, etc. In recent years, graph machine learning~(GML) models, initiated from DeepWalk~\cite{perozzi2014deepwalk} until recent graph neural networks (GNNs)~\cite{GCN,hamilton2017inductive,velivckovic2017graph}, have achieved superior performance on graph-based tasks such as node classification~\cite{GNNsurvey,2017Graph}, link prediction~\cite{linksurvey} and graph classification~\cite{waikhom2021graph,zhou2021graph},
and thus be extensively applied on social network analysis~\cite{tan2019deep}, recommendation systems~\cite{ying2018graph,wang2020multi}, knowledge representation~\cite{nathani2019learning,sun2020knowledge} and protein function prediction~\cite{shen2021npi}, etc.

\begin{figure}[htbp] 
	\centering
	\setlength{\abovecaptionskip}{0.cm}
	\subfigure[Node degree distribution]
	{\label{degree}\includegraphics[width=0.45\linewidth]{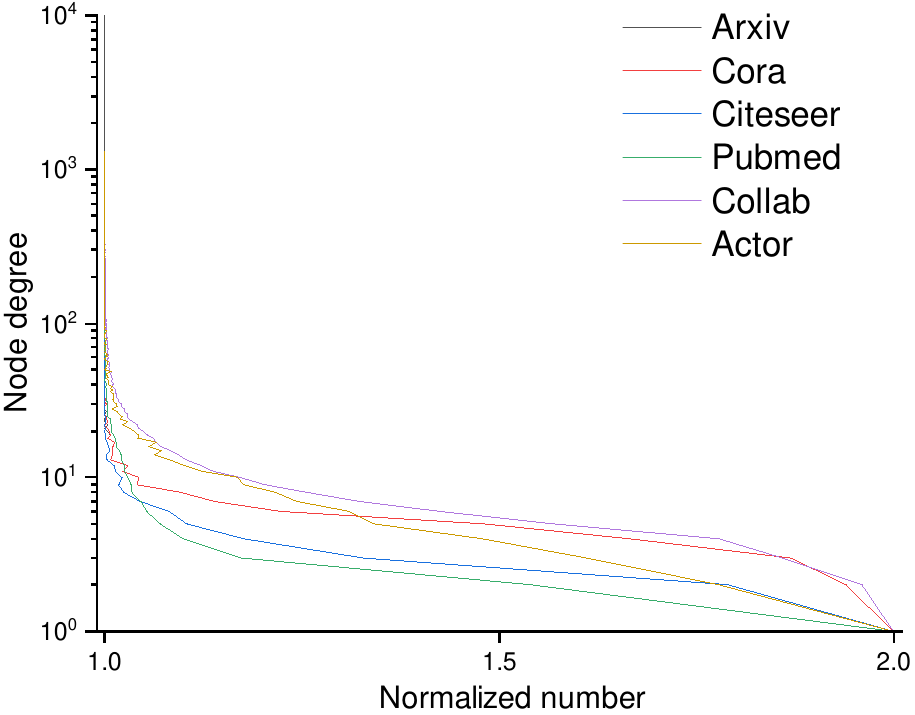}}
	\subfigure[Node degree distribution of new edges]
	{\label{collab}\includegraphics[width=0.45\linewidth]{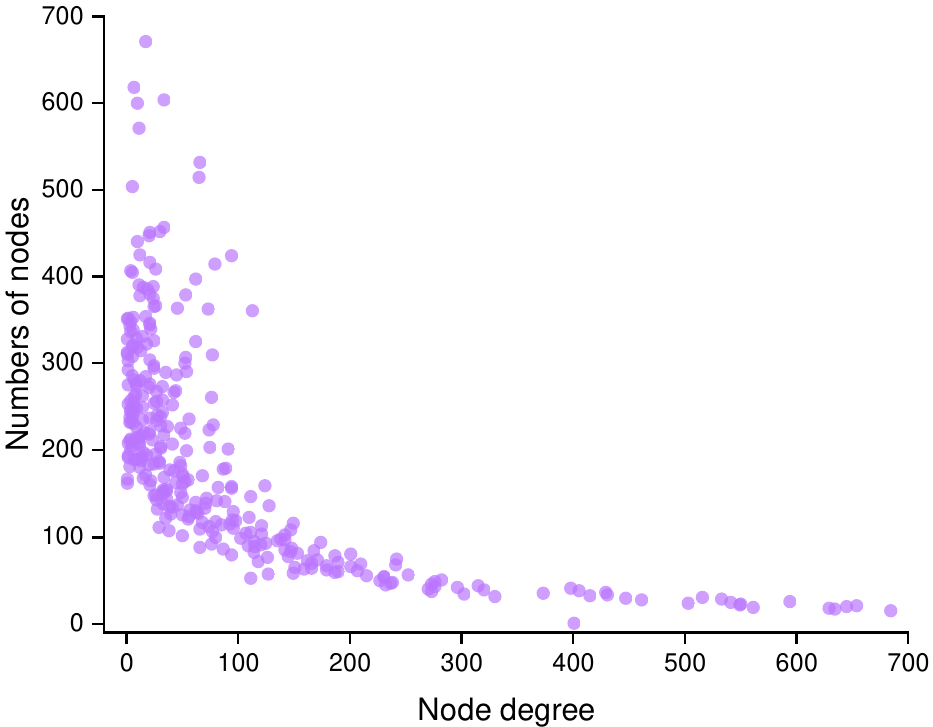}}
	\caption{The \textbf{left} figure shows the long-tail distribution of node degrees on ﬁve datasets, where the number of nodes is normalized to avoid the effect of  dataset scale, and the \textbf{right} figure shows the node degree distribution of emerging edges on the Collab dataset.}
\end{figure}

Despite the success of existing GML approaches, the literatures have rarely noticed the structural imbalance widely existent in graphs. Generally, only a few nodes in the graph have more important structure and higher inﬂuence. 
Taking the node degree as an example, it follows the power law distribution in many  scale-free networks~\cite{barabasi2003scale} (Figure~\ref{degree}). The Pagerank~\cite{page1999pagerank} value is another indicator that can better reﬂect the importance of nodes. In this paper, nodes with high structural importance are called \emph{hub nodes} and nodes with low structural importance are called \emph{tail nodes}.

\vspace{-1mm}

This paper argues that the existence of structural imbalance may compromise the robustness of current GML methods due to the learning bias in the training process. In DeepWalk~\cite{perozzi2014deepwalk} and other random walk-based models, the sampling amount of tail nodes is far less than that of hub nodes as the number of random walks is proportional to the degree of nodes. Besides, the model scalability is limited by the redundancy in random walks. In GNNs, the message passing mechanism enables nodes to aggregate information from their neighborhoods~\cite{GCN,hamilton2017inductive,velivckovic2017graph,chen2020simple}. However, information is more difficult to propagate to tail nodes resulting in the performance degradation of the model in downstream tasks. Figure \ref{collab} depicts the the node degree distribution of newly emerging edges over time on the Collab dataset, demonstrating that many tail nodes or non-hub nodes tend to have more new neighbors even than the hub nodes have in the future. As a result, enhancing the learning of tail nodes will help to predict more accurately links.

A possible solution to this structual imbalance is to increase the receiving field of tail nodes by stacking GNN layers, but at the cost of over-smoothness~\cite{li2018deeper} due to the rapid expansion of neighborhoods. This paper attempts to tackle this issue from a new perspective, introducing graph augmentation learning to selectively enhance the local structures of nodes according to their structural importance to achieve a more reliable GML model. For hub nodes, although they have a considerable number of neighbors, some of them may not be similar or relevant and will generate noise during aggregation. Taking Twitter as an example, celebrities often have a large number of followers because of their prestige and popularity. But most of them have no similarities with the celebrities they follow. For tail nodes, although they have few neighbors at present, they might be cold-start users with many latent neighbors to explore. 

In this paper, a structural imbalance aware graph augmentation method is proposed. Firstly, hub nodes and tail nodes are identified via a Pagerank-based sampling strategy. Secondly, a selective augmentation method is proposed to remove the noise neighbors for hubs on the one hand, and explore the latent neighbors for tails on the other hand. Subsequently, a generative model~(e.g. GAN~\cite{goodfellow2014generative} or VAE~\cite{kingma2013auto}) is utilized to further generate pseudo neighbors for tails. Finally, after alleviating the structural imbalance between hubs and tails, a more robust GML model can be obtained.

Our main contributions are summarized as follows:
\begin{itemize}
	\item[$\bullet$] To the best of our knowledge, this is the first to focus on the structural imbalance problem encountered by GML models, which is essentially different from the existing research on node label imbalance.
	\item[$\bullet$] We proposes a selective graph augmentation method to better learn the node representation, especially for tail nodes and to alleviate the structural imbalance as well.
	\item[$\bullet$] Our method can be applied on a variety of GML models with convenience. Extensive experimental results show that our method achieves superior performance on benchmark datasets.
\end{itemize}

\section{Related Work}
\subsection{Hub/Tail Aware GML\label{scalefree}} 
The existing GML methods range from non-deep learning methods such as matrix factorization~\cite{cai2010graph} and DeepWalk~\cite{perozzi2014deepwalk} to deep graph learning methods such as GNNs~\cite{GCN,hamilton2017inductive,velivckovic2017graph}. 

To our best knowledge, only limited works treat hub nodes and tail nodes differently. TailGNN~\cite{liu2021tail} proposes to transfer the rich neighborhood information of head nodes (i.e., hub nodes) to tail nodes based on the learned neighborhood translation vector. But it may also transfer noise data such as low relevant or less important neighbors. ColdBrew~\cite{zheng2021cold} utilizes a teacher-student distillation model to solve the extremely cold start problem of GNNs. The MLP is trained as the student model to simulate node embeddings learned by GNNs (the teacher model) in order to find latent neighbors for tail nodes and isolated nodes. CenGCN~\cite{xia2022cengcn} transforms the graph by redistributing the weights beetween hubs and their neighbors based on similarity and centality indices, and assigns new weights to non-hub nodes which share the same hub neighbor based on a non-hub attention mechanism. Since it uses label propagation to calculate the similarity between hubs and tails, the quality of similarity caculation depends a lot on the labels of nodes.

\subsection{Graph Structural Augmentation}
Recently, graph data augmentation research has been actively conducted to handle data noise and sparsity issues in graph learning~\cite{ding2022data,zhao2022}. Graphs are intrinsically relational where the structures are critical to graph analysis and inference. Hence, many efforts are put on graph structural augmentation. The methods in the literatures mainly include edge-level augmentation and node-level augmentation~\cite{zhao:graphsmote,you2021graph}.

In this paper, we mainly focus on the edge-level augmentation. AdaEdge~\cite{chen2020measuring} iteratively adds edges between nodes which have the same predicted labels with high confidence in the modified graph. DropEdge~\cite{rong2019dropedge} randomly removes a certain number of edges from the input graph at each training epoch, to avoid over-fitting and over-smoothing. PTDNet~\cite{luo2021learning} proposes a parameterized topological denoising network, to improve the robustness and generalization performance of GNNs by learning to drop task-irrelevant edges. GAug~\cite{zhao2021data} utilizes edge predictors to encode graph homophilic structure, and thus to predict missing links and links that should have been removed.

This paper attempts to enhance the structure of both hub nodes and tail nodes from the perspective of graph augmentation to reduce the structural imbalance and improve the robustness of GNN.

\section{Preliminaries}
This section introduces the preliminaries of our work, including formalized definitions and problem descriptions.
\subsection{Definitions}
\paragraph{Graphs.}
Given a graph $\mathcal{G}=(\mathcal{V},\mathcal{E}, \mathcal{A},\mathcal{X}, \mathcal{Y})$, $\mathcal{V}$ denotes the set of nodes, $\mathcal{E}$ denotes the set of edges, $e_{ij}\in \mathcal{E}, i,j\in \mathcal{V}$ is the edge between node $i$ and $j$, $\mathcal{A}\in\mathbb{R}^{\left| \mathcal{V} \right| \times \left| \mathcal{V} \right|}$ denotes the adjacent matrix, $\mathcal{X} \in \mathbb{R} ^{\left| \mathcal{V} \right|\times d_{\mathcal{X}}} $ denotes the attribute matrix where $d_{\mathcal{X}}$ is the dimension of nodes' attributes, and  $\mathcal{Y}\in \mathbb{R}^{\left| \mathcal{V} \right|}$ represents the node label vector, which could be provided depending on the specific downstream tasks. 

\paragraph{Hub and Tail nodes\label{hubtail}}
As mentioned above, this paper uses the \emph{Pagerank} value to sample hub and tail nodes as it can represent the centrality and importance of nodes on the graph more effectively than the node degree. Formally, the set of hub nodes is represented as: $\mathcal{V} _{hub}=\left\{ v: PR(v)\geq \gamma \right\} $, and the set of tail nodes is represented as: $\mathcal{V} _{tail}=\left\{ v: PR(v)\leq \zeta \right\}$, where $PR(v)$ indicates the Pagerank value of node $v$, which can be caculated by Eq.~\ref{pagerank}, 
\begin{equation}
	PR\left( v_i \right) =d\left( \sum_{v_j\in In\left( v_i \right)}{\frac{PR\left( v_j \right)}{\left|\mathcal{N}\left( v_j \right)\right|}} \right) +\frac{1-\xi}{n}, i=1,2,...,n \label{pagerank}
\end{equation}
Here, $In(v_i)$ denotes the set of nodes pointing to $v_i$, $\left|\mathcal{N}\left( v_j \right)\right|$ indicates the number of neighbors of $v_j$, $\xi$ is the damping coefficient which is set to 0.85 by default. For an undirected graph, it is generally transformed into a directed graph, where each undirected edge corresponds to two directed edges. However, it is difficult to determine the value of $\gamma$ and $\zeta$. So a sampling strategy based on Pagerank will be adopted to split hub nodes and tail nodes, which is described in detail in Section \ref{sample}. 
\subsection{Problem Description}
Our work aims to find a structural augmentation mapping function $f_\theta:\mathcal{G}\rightarrow \tilde{\mathcal{G}}$. Here, $\tilde{\mathcal{G}} = (\tilde{A}, \tilde{X})$ is the augmented graph, which has a more balanced structure than the original graph $\mathcal{G}$ and moreover the low-dimensional embedding $\mathcal{H}$ learned from $\tilde{\mathcal{G}}$ has higher robustness in downstream tasks. 

As milestone GML methods, graph neural networks (GNNs) learn node representations by aggregating structure and attribute information through the message passing mechanism~\cite{zhou2021graph}. Specifically, The $l$-th layer of a GNN aggregates the neighborhood information for node $v$ and combines it with the self-representation of $v$ to obtain the representation vector $h_v$:
\begin{equation}
	h_{v}^{(l)}=f\left( h_{v}^{(l-1)},\left\{ h_{u}^{(l-1)}:u\in \mathcal{N} _v \right\} ;\Theta ^{(l)} \right) \label{GNN}
\end{equation}
where $f(\cdot)$ indicates the message passing function with trainable parameter $\Theta^{(l)}$ and $h_v^{(0)}$ in the first layer is initialized with $\mathcal{X}_v$. 

\section{Proposed Method}
\begin{figure*}[!t]
	\centering
	\includegraphics[height=2.5in]{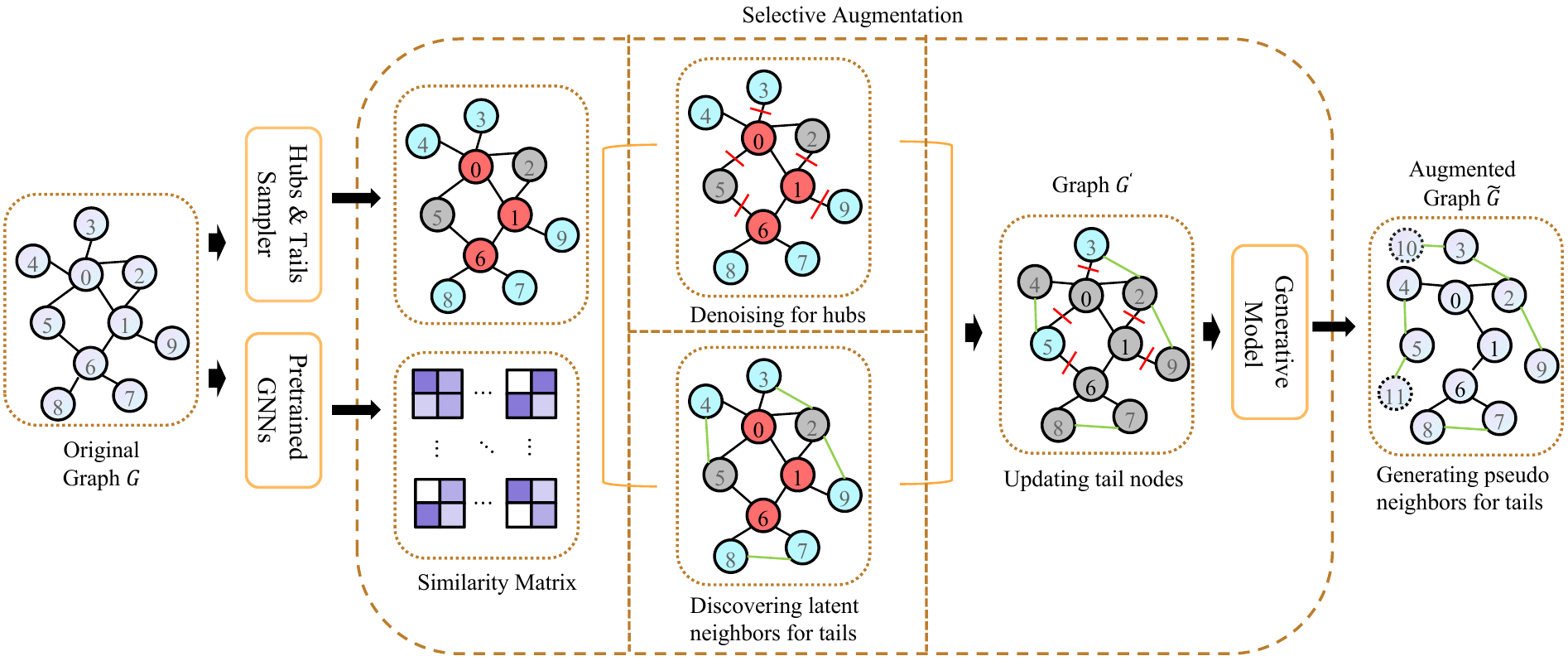}
	\caption{Given the input graph $\mathcal{G}$, nodes 0, 1, 6 are sampled as hub nodes and marked in red, while nodes 3, 4, 7, 8, 9 are sampled as tail nodes and marked in light blue. After obtaining the pair wise node similarity matrix based on node embeddings learned by the pretrained GNNs, hubs and tails are augmented, respectively. In the augmented graph $\mathcal{G}^{'}$, tail nodes are then re-sampled (i.e. nodes 3, 5), and the generative model is further utilized to generate pseudo neighbors (i.e., node 10, 11) for tail nodes in $\mathcal{G}^{'}$. The augmented graph $\tilde{\mathcal{G}}$ is ﬁnally obtained.}
	\label{framework}
\end{figure*}

The proposed method SAug (selective augmentation for structural imbalance graphs) is composed of three steps: Firstly, hub nodes and tail nodes are sampled through a Pagerank-based strategy; Secondly, two different edge augmentation strategies are designed for hub nodes and tail nodes, respectively; Finally, the neighborhood is further augmented for tail nodes by generating pseudo neighbors. The framework of SAug is shown in Figure~\ref{framework}.

\subsection{Sampling Hub and Tail Nodes \label{sample}}
As mentioned in Section \ref{hubtail}, the Pagerank value is used as the criterion to identify hub nodes and tail nodes. According to Eq.~\eqref{pagerank}, the Pagerank value of each node has the following properties normalized by smoothing term $\cfrac{1-d}{n}$: 
\begin{gather}
	PR\left( v_i \right) > 0,i=1,2,...,n
	\\
	\sum_{i=1}^n{PR\left( v_i \right) = 1}
\end{gather}

Therefore, $PR(v_i)$ can be directly used as the sampling probability. For a given node $v_i$, the higher the $PR(v_i)$, the more likely it is to be sampled as a hub node and the lower the $PR(v_i)$, the more likely it is to be sampled as a tail node. Due to the centrality of hub nodes, the Pagerank values of hubs are much higher than those of non-hubs. In this paper, the nodes with Pagerank value larger than $K$ times the average $PR_{avg}$ are sampled as the hubs to form the $\mathcal{V}_{hub}$ set, denoted as: $PR(v) \ge K*PR_{avg}, v\in \mathcal{V}_{hub}$. Here $K$ is initialized with 2. Then, $M\%$ of the remaining nodes are sampled to form the $\mathcal{V}_{tail}$ set, where $M$ is initialized with 30. 

\subsection{Selective Structural Augmentation}
This paper adopts different augmentation strategies for hub nodes and tail nodes considering the structural imbalance of graphs. For hub nodes, the noise edges are removed. For tail nodes, the latent neighbor in the whole graph are explored to enrich their local structure. 

\paragraph{Pretraining GNNs.}
Firstly, we need to pre-train GML models on the entire graph to determine the noise edges and latent edges based on the pretrained node embeddings. 

Most existing GNNs are based on the homophilic hypothesis or homodirectional preference hypothesis, that is, nodes with the same label or similar features are more inclined to establish mutual connections. Therefore, we use GNNs to encode the graph and calculate the similarity between node pairs. Any GNN model (such as GCN, GraphSAGE, GAT, etc.) can be selected for pre-training. Additionally, the label information is also used as the weight matrix for similarity calculation to further improve the method. 

In our method, two GNNs models are pretrained respectively as the link predictor and the label classifier to obtain the label-dependent and link-dependent embeddings which are the basis for the augmentation operation below.  
The embeddings obtained by these two models are represented as $Z_{link} \in \mathbb{R}^{\left| \mathcal{V} \right|\times d_{out}}$ and $Z_{label} \in \mathbb{R}^{\left| \mathcal{V} \right|\times \mathcal{Y}_{num}}$, respectively, where $d_{out}$ is the dimension of the output layer of the link predictor, and $\mathcal{Y}_{num}$ is the number of labels in a given dataset. 

In the link predictor, the negative samples are from unconnected node pairs to form $\mathcal{E}_{neg}$, where $\left|\mathcal{E}_{neg}\right|=\left|\mathcal{E}\right|$. If $e_{ij}\in \mathcal{E}$, the label $\mathcal{Y}^{e_{ij}}=1$ and if $e_{ij}\in \mathcal{E}_{neg}$, the label $\mathcal{Y}^{e_{ij}}=0$. Therefore, the loss function of the link predictor could be expressed as:
\begin{equation}
	\begin{split}
		&\mathcal{L} _{lp}=\sum_{e_{ij}\in \left( \mathcal{E} \cup \mathcal{E}_{neg} \right)}{\mathrm{BCE}\left( \sigma \left( z_{link}^{i}\cdot{z_{link}^{j}}^{\top} \right) , \mathcal{Y}^{e_{ij}} \right)} 
		\\
		&+ \lambda \left\| \Theta _{lp} \right\| _{2}^{2}, \label{lp}
	\end{split}
\end{equation}
where BCE is the binary cross-entropy function, $z_{link}^i = Z_{link}[i, :]$, and $\Theta_{lp}$ is the set of model parameters with regularization coefficient $\lambda$.

The loss function of the label classifier is:
\begin{gather}
	\mathcal{L} _{nc}=\mathrm{CE}\left( \hat{\mathcal{Y}}, \mathcal{Y} \right) + \mu \left\| \Theta _{nc} \right\| _{2}^{2} \label{nc}
	\\
	\hat{\mathcal{Y}}=argmax\left( \sigma \left( Z_{label} \right) \right) 
\end{gather}
where $\hat{\mathcal{Y}}$ is the label probability predicted by the pretrained node classification model, and CE is the cross-entry function.

After obtaining the embedding matrix $Z_{link}$ and $Z_{label}$, we conduct different augmentation strategy for hub nodes and tail nodes, respectively.
\paragraph{Denoising irrelevant neighbors for hubs.\label{denoise}} 
The hub node and its irrelevant neighbor nodes usually have different labels and features. 

The similarity matrix between the hub node $v_i$ and its neighbors $\mathcal{N}(v_i)$ is calculated as:
\begin{equation}
	\begin{split}
		S_{hub}^{i}=\left( z_{label}^{i}\cdot {z_{label}^{\mathcal{N} \left(i \right)}}^{\top} \right) \odot \left( z_{link}^{i}\cdot {z_{link}^{\mathcal{N} \left( i \right)}}^{\top} \right),
		\\
		{S_{hub}^{i}}^{\top}\in \mathbb{R} ^{\left| \mathcal{N} \left( i \right) \right|\times 1} \label{hub_sim}
	\end{split}
\end{equation}
where $\cdot$ is the inner product operation, and $\odot$ is the \emph{Hadamard} product operation. The similarity score between node pairs with different labels will be reduced with the introduction of $Z_{label}$, which enables us to better distinguish noise neighbors. The neighbor nodes with similarity score less than $L$ are marked as noise neighbors and the corresponding edges are removed where the hyperparameter $L \in (0,1)$ is initialized with 0.1.

\paragraph{Discovering latent neighbors for tails.\label{discover}}
Similarly, we utilize $Z_{link}$ and $Z_{label}$ to calculate the similarity between the tail node and the remaining nodes to find its latent neighbors. The similarity matrix is calculated as:
\begin{equation}
	\begin{split}
		S_{tail}=\left( z_{label}^{\mathcal{V} _{tail}}\cdot {Z_{label}}^{\top} \right) \odot \left( z_{link}^{\mathcal{V} _{tail}}\cdot {Z_{link}}^{\top} \right), 
		\\
		S_{tail}\in \mathbb{R} ^{\left| \mathcal{V} _{tail} \right|\times \left| \mathcal{V} \right|} \label{tail_sim}
	\end{split}
\end{equation}
Given a tail node $v_i$, two different neighbor selection strategies are used: i) Similar to the denoising operation for hub nodes, a threshold $P\in (0,1)$ is picked. The node $v_j$ will be selected as the latent neighbor if $S^{ij}_{tail} >= P$. ii) The nodes having the top $Q$ similarity scores in $S_{tail}^{i}$ are marked as latent neighbors and added to $\mathcal{N}(v_i)$.

After the above operation, the graph $\mathcal{G}$ is augmented to $\mathcal{G}^{'}=(\mathcal{V},\mathcal{E}^{'}, \mathcal{A}^{'},\mathcal{X}, \mathcal{Y})$, where $\mathcal{E}^{'}$ is the  modified edge set after dropping and adding edges, and $\mathcal{A}^{'}$ is the modified adjacent matrix.

\paragraph{Generating pseudo neighbors for tails.}
The above augmentation operations change the graph's structure, so the tail nodes need to be re-sampled for subsequent operations. We utilize a generative model to generate pseudo neighbor nodes for tail nodes to further enrich their local structures. Considering the high similarity between the features of tail nodes and their local neighborhood, we directly feed the features of neighbor nodes similar to tail nodes into the generative model. 

The similarity between a tail node $v_i$ and its neighbors is first calculated by the cosine measurement based on the adjacency matrix $\mathcal{A}^{'}$ obtained above, which is represented as: 

\begin{equation}
	S_{cosine}^{i}=\frac{\mathcal{X}^{i}\cdot {\mathcal{X}^{\mathcal{N}(i)}}^{\top}}{\left\| \mathcal{X}^{i} \right\| \times \left\| \mathcal{X}^{\mathcal{N}(i)} \right\|}, v_i\in \mathcal{V} _{tail}
\end{equation}
for node $v_i$, its most similar neighbor $v_j$ is picked out, i.e. $v_j=argmax(S_{cosine}^{i})$. 
For all nodes in $\mathcal{V}_{tail}$, we obtain a similar neighbor set $\mathcal{V}_{sim}$ and construct its corresponding feature matrix $\mathcal{X}_{\mathcal{V}_{sim}}$ and labels $\mathcal{Y}_{\mathcal{V}_{sim}}$ for generative operations.

A fully connected neural network is used as the generator G, to learn a mapping function $f:\mathcal{Z} \rightarrow \mathcal{X}$. Specifically, the noise $z\in\mathbb{R}^{d_z}$ is fed into the generator G, which maps $z$ into node feature space $\mathcal{X}$.
In this paper, one pseudo neighbor node is generated for one tail node. However, we do not explore the relationship between the pseudo nodes and the real nodes in the graph since it could introduce extraneous noise. The loss function to train G is: 
\begin{equation}
	\mathcal{L} _{gen}=\sum_{v_i\in \mathcal{V}_{sim}}{\left\| x_{gen}^{i}-x_{sim}^{i} \right\| _{2}^{2}}+\alpha \left\| \Theta _{gen} \right\| _{2}^{2}
\end{equation}
where $x_{gen}^{i}$ means the $\mathcal{X}_{gen}[i, :]$, $\Theta_{gen}$ is the set of weights in the generator with regularization coefficient $\alpha$. The labels $\mathcal{Y}_{\mathcal{V}_{sim}}$ of $\mathcal{V}_{sim}$ are assigned to the corresponding generated nodes $\mathcal{V}_{gen}$ which are prepared for the loss function. Finally, we get the modified graph $\tilde{\mathcal{G}}=(\tilde{\mathcal{V}}, \tilde{\mathcal{E}}, \tilde{\mathcal{A}}, \tilde{\mathcal{X}}, \tilde{\mathcal{Y}}, \mathcal{C})$, where $\tilde{\mathcal{V}}$ is the updated node set which includes the generated nodes $\mathcal{V}_{gen}$, $\tilde{\mathcal{E}}$ represents the updated edge set which contains new edges between $\mathcal{V}_{gen}$ and $\mathcal{V}_{tail}$, $\tilde{\mathcal{A}}$ is the adjacency matrix updated based on $\tilde{\mathcal{A}}$, $\tilde{\mathcal{X}}$ and $\tilde{\mathcal{Y}}$ represent the features and labels of $\tilde{\mathcal{V}}$, respectively. $\mathcal{C}=\{0,1\}$ indicates whether node $v_i$ is a pseudo node or not, if $v_i$ is generated, $\mathcal{C}_{i}=0$, vice versa.

A two-layer GCN model is used as the discriminator D, with the input of the graph $\tilde{\mathcal{G}}$ obtained above. The target of D is to discriminate whether node $v_i$ is generated by G or not, and to classifyof node $v_i$. Here, GCN is actually treated as a node classifier. The loss function of the discriminator is: 
\begin{equation}
	\mathcal{L} _{dis}=\mathrm{CE}\left( \hat{\tilde{\mathcal{Y}}},\tilde{\mathcal{Y}} \right) +\mathrm{BCE}\left( \hat{\mathcal{C}},\mathcal{C} \right) + \beta \left\| \Theta _{dis} \right\| _{2}^{2}
\end{equation}
where $\hat{\tilde{\mathcal{Y}}}$ and $\hat{\mathcal{C}}$ are both maximum terms in the output of discriminator and the last term is the regularization term.

Finally, the objective function of the generative module is summerized as:
\begin{equation}
	\begin{split}
		\underset{\mathrm{G}}{\min}\underset{\mathrm{D}}{\max}V\left( \mathrm{D},\mathrm{G} \right) =\mathbb{E} _{x\sim p_{\mathcal{X}_{sim}}\left( x \right)}\left[ \log \mathrm{D}\left( x \right) +\mathcal{L} _{dis} \right] \label{GANloss}
		\\
		+\mathbb{E} _{z\sim p_z\left( z \right)}\left[ \log \left( 1-\mathrm{D}\left( \mathrm{G}\left( z \right) \right) \right) +\mathcal{L} _{gen} \right] 
	\end{split}
\end{equation}

\section{Experiments}
In this section, SAug is compare to several baselines methods via downstream tasks of node classification and link prediction. We also conduct the ablation study for SAug and discuss its sensitivity to various hyperparameters. 
\subsection{Experiment Setup}
\paragraph{Datasets.} Five benchmark datasets are used in this paper including Cora~\cite{mccallum2000automating}, Citeseer~\cite{giles1998citeseer}, Chameleon~\cite{pei2020geom}, Squirrel~\cite{rozemberczki2021multi}, and Actor~\cite{tang2009social}. Statistics of each dataset and more details are summarized in Appendix~\ref{datasets_details}.

\paragraph{Comparison methods.}
We compare our model with three categories of baseline methods: i) Base GNNs, including GCN~\cite{GCN}, GraphSAGE~\cite{hamilton2017inductive} and GAT~\cite{velivckovic2017graph}, which are also the backbone model in our method. 
ii) Hub/tail aware GNNs, including TailGNN~\cite{liu2021tail}, ColdBrew~\cite{zheng2021cold} and CenGCN~\cite{xia2022cengcn}, which are discribed in Sec~\ref{scalefree}. For fair comparison, the split criteria of hubs and tails in comparative models is consistent with ours, that is, based on Pagerank values. iii) Graph structural augmentation models, including  DropEdge~\cite{rong2019dropedge}, AdaEdge~\cite{chen2020measuring}, PTDNet~\cite{luo2021learning}, and GAug~\cite{zhao2021data}. 

\subsection{Tail Node Classification}
First of all, the tail node classiﬁcation task is experimented to verify the robustness of SAug on tail nodes. Specifically, 30\% nodes are first sampled as tail nodes according to the Pagerank-based sampling strategy, which are divided into validation set and testing set with a ratio of 2:1, and the remaining nodes are utilized as the training set. All augmentation operations in SAug are performed in the training set. Moreover, the semi-supervised learning setting is adopted, where the labels of 10 nodes per class in the training set are regarded as the ground-truth in each training epoch. All augmentation operations in SAug are performed in the training set.

For a detailed comparison, GCN is first adopted as the backbone model for all GNN-based baselines. Then, GraphSAGE and GAT are complementarily adopted to verify the performance of SAug on other GNN backbones. The Macro-F1 and the Micro-F1 score are used as the evaluation indicator of the node classification task. 

\begin{table*}[htbp] 
	\centering
	\resizebox{\textwidth}{!}{
		\begin{tabular}{l*{10}{r}}
			\toprule
			\multirow{2}*{Methods} &\multicolumn{2}{c}{Cora} &\multicolumn{2}{c}{Citeseer} &\multicolumn{2}{c}{Chameleon} &\multicolumn{2}{c}{Squirrel} &\multicolumn{2}{c}{Actor}
			\\
			\cmidrule(lr){2-3}\cmidrule(lr){4-5}\cmidrule(lr){6-7}\cmidrule(lr){8-9}\cmidrule(lr){10-11}
			&Macro-F1 &Micro-F1 &Macro-F1 &Micro-F1 &Macro-F1 &Micro-F1 &Macro-F1 &Micro-F1 &Macro-F1 &Micro-F1
			\\
			\midrule
			GCN &84.9±0.6 &86.8±0.7 &69.3±0.5 &74.2±0.5 &45.0±1.1 &47.8±1.3 &26.1±1.7 &27.6±1.5 &18.9±0.4 &27.7±0.5
			\\
			GCN w/ $Gen$ &85.3±0.5 &87.0±0.6 &70.1±0.7 &74.9±0.6 &45.2±2.1 &48.1±2.0 &24.7±2.5 &26.6±2.0 &19.1±0.5 &27.9±0.6
			\\
			\midrule 
			DropEdge &85.1±0.5 &87.2±0.4 &68.7±0.7 &73.6±0.6 &45.4±1.7 &48.1±1.4 &25.4±2.4 &27.3±2.6 &19.4±1.6 &27.5±1.4
			\\
			AdaEdge &85.6±0.7 &87.4±0.8 &71.7±0.6 &75.6±0.8
			&45.7±1.1 &48.4±2.1 &26.9±2.3 &28.1±2.5 &20.1±0.6 &29.2±0.7
			\\
			PTDNet &85.9±0.4 &87.3±0.6 &71.4±0.4 &75.9±0.6
			&46.4±1.3 &49.3±1.6 &27.4±2.1 &28.5±2.9 &20.8±1.2 &29.7±1.3
			\\
			GAug &86.9±0.4 &\underline{88.6±0.7} &73.6±0.7 &77.1±0.9 &51.8±0.9 &53.5±0.9 &31.1±2.1 &32.3±1.4 &23.4±1.0 &31.6±1.5
			\\
			\midrule
			ColdBrew &85.2±0.5 &87.0±0.4 &68.9±0.6 &74.3±0.5 &47.9±1.4 &49.7±1.2 &27.7±1.4 &29.2±1.9 &20.3±1.4 &29.0±1.2
			\\
			%DP &68.3±0.4 &71.9±0.6 &56.4±0.5 &57.1±0.8 &39.1±1.6 &41.7±1.3 &28.6±2.1 &30.4±2.3 &13.7±1.5 &19.9±2.1
			TailGCN &86.8±0.5 &88.4±0.6 &73.6±0.4 &77.5±0.8 &\underline{52.3±1.4} &\underline{53.3±1.2} &\underline{32.7±1.2} &\underline{33.8±0.8} &\underline{34.7±1.6} &\underline{42.5±1.5}
			\\
			CenGCN &\underline{87.1±0.5} &88.1±0.7 &\underline{74.1±0.3} &\underline{78.2±0.2}
			&50.9±1.3 &52.7±1.1 &30.4±2.6 &31.7±2.1 &32.5±1.6 &39.1±1.3
			\\
			\midrule
			Denoise &85.9±0.6 &87.6±0.6 &69.5±0.4 &74.3±0.5 &45.5±1.7 &48.1±1.9 &25.0±1.9 &27.1±1.6 &19.3±0.6 &27.9±0.5
			\\
			SAug$_{thr}$ w/o $Gen$ &87.8±0.7 &\textbf{89.4±0.5} &74.2±0.9 &79.6±0.8 &51.6±1.4 &53.7±1.6 &\textbf{34.6±2.1} &\textbf{35.3±1.5} &37.8±1.1 &49.4±0.6
			\\
			SAug$_{thr}$ &\textbf{88.0±0.8} &89.4±0.8 &74.6±1.2 &\textbf{79.8±0.8} &52.4±1.1 &54.7±1.2 &33.7±2.8 &34.4±2.4 &\textbf{38.1±1.6} &\textbf{49.6±0.5} 
			\\
			SAug$_{top}$ w/o $Gen$ &86.2±0.6 &87.6±0.5 &74.9±0.8 &79.6±1.5 &\textbf{53.0±0.8} &\textbf{54.8±0.8} &29.7±1.9 &30.7±1.5 &28.6±1.0 &40.6±0.7
			\\
			SAug$_{top}$ &86.9±0.7 &87.8±0.6 &\textbf{75.6±0.9} &79.7±0.8 &52.6±1.8 &54.7±2.0 &28.3±1.7 &30.1±2.1 &28.1±1.5 &40.4±0.6
			\\
			\bottomrule
		\end{tabular}
	}
	\caption{Performance on tail node classification with GCN backbone. The bolded data in each column is the best result, and the underlined data are the best results in comparative models. Henceforth, all tabular results are in percent.}
	\label{gcn_nc}
\end{table*}

\paragraph{GCN-based models.} The results of tail node classification are shown in Table \ref{gcn_nc}, where $Gen$ and $Denoise$ denote the generative module and the denoising module in SAug, respectively. SAug$_{thr}$ is the SAug variant using the threshold operation, and SAug$_{top}$ is the SAug variant using the topQ operation. 

Overall, SAug series achieve the best performance in tail node classification on five benchmark datasets. Among four graph structural augmentation models (DropEdge, AdaEdge, PTDNet and GAug), DropEdge does not significantly improve the performance of the base model due to its random edge dropping strategy; AdaEdge iteratively adds/removes edges between node pairs whose labels are predicted to be same/different in each training epoch, but this approach performs only a limit better than DropEdge probably because it may propogate error labels; PTDNet achieves limited performance improvement, since its denoising operation is applied on all nodes equally, which might dilute the neighborhood of tail nodes; GAug uses GAE as the edge predictor to add or drop edges, but it also adopts a consistent augmentation strategy for both hubs and tails without taking the structural imbalance into consideration.

Among hub/tail aware methods, ColdBrew utilizes MLP as the student model to find latent neighbors for tail and isolated nodes. However, MLP cannot utilize the neighborhood information of tail nodes, which may explain ColdBrew's low performance improvement in tail node classification; TailGNN performs much better than ColdBrew because it focuses more on improving the representation of tails, but it may transfer the noise information from the neighborhood of hubs to tails, so it performs worse than SAug; CenGCN may calculate inaccurate weight calculation due to the absence of labels, so the performance of CenGCN is not very stable, dependentmore on the label quality of the dataset. Different from TailGNN and CenGCN, SAug removes noise neighbors for hubs and meanwhile obtains more accurate embeddings for the unlabeled nodes owing to pre-training GNNs under semi-supervised learning framework. These may explain why SAug outperforms other hub/tail aware models.

\paragraph{Other GNNs-based model.} We also use GraphSAGE and GAT as the backbone to replace GNN for GAug, TailGCN and SAug. Note that CenGCN is proposed on the basis of GCN and cannot be modified to the version of other GNNs. Part of the comparison results are summarized in Table 2. See Appendix~\ref{additional_experiments} for more details.
\begin{table}[h]
	\centering
	\resizebox{\linewidth}{!}{
	\begin{tabular}{lrrrrr}
		\toprule
		Methods &Cora &Citeseer &Chameleon &Squirrel &Actor
		\\
		\midrule
		GraphSAGE &87.0±0.8 &74.2±0.5 &47.0±1.0 &32.9±1.6 &42.4±1.0
		\\
		\midrule
		GAugSAGE  &\underline{92.6±0.5} &\underline{79.0±0.6} &53.1±1.3 &40.4±1.9 &53.9±1.6
		\\
		TailSAGE &92.1±0.6 &78.5±0.8 &\underline{55.9±1.4} &\underline{43.7±2.3} &\underline{60.4±1.5}
		\\
		\midrule
		SAugSAGE &\textbf{94.0±0.4} &\textbf{81.8±0.7} &\textbf{61.3±1.5} &\textbf{56.6±0.9} &\textbf{64.9±1.2}
		\\
		\midrule
		\midrule
		GAT &86.1±0.9 &75.5±1.1 &45.9±1.9 &30.2±2.9 &29.4±0.9
		\\
		\midrule
		GAugGAT &\underline{87.0±0.7} &\underline{77.1±0.6} &53.7±2.1 &31.4±2.6 &29.6±1.1
		\\
		TailGAT &86.9±0.5 &76.4±0.8 &\underline{55.1±2.0} &\textbf{32.5±2.4} &\textbf{30.7±1.7}
		\\
		\midrule
		SAugGAT &\textbf{87.4±0.7} &\textbf{77.9±0.8} &\textbf{56.1±1.9} &\underline{32.0±2.8} &\underline{30.2±1.3}
		\\
		\bottomrule
	\end{tabular}
	}
	\caption{Micro-F1 scores on tail node classiﬁcation with other GNN variants. In each comparison group, the best result is bolded and the second best result is underlined.}
	\label{gnn_variants}
\end{table}

Our method can effectively enhance base GNNs to varying degrees and overall achieve the optimal performance compared to the corresponding GAug and TailGNN. In the Squirrel and Actor datasets, the performance of SAug-GAT is slightly lower than that of TailGAT. We argue that $Z_{label}$ and $Z_{link}$ of Squirrel and Actor obtained from the pretrained GAT are not as accurate as GCN and GraphSAGE, which results in low similarity scores between node pairs. It might also explain why SAugGAT performs worse than TailGAT on Squirrel and Actor.

\subsection{Overall Node Classification} 
This section further investigates the overall node classification performance on the entire graph. We conduct experiments on five datasets under standard split criterion~\cite{GCN,pei2020geom}. The results in Table~\ref{overall_nc} show that our model achieves considerable improvement over backbone GNN models again. 

\begin{table}[h]
	\centering
	\resizebox{\linewidth}{!}{
		\begin{tabular}{lrrrrr}
			\toprule
			Methods &Cora &Citeseer &Chameleon &Squirrel &Actor 
			\\
			\midrule
			GCN &80.1±0.6 &63.1±0.9 &35.4±1.1 &23.1±1.2 &23.9±1.0
			\\
			GAug &83.6±0.5 &\underline{69.3±0.7} &37.1±1.3 &\underline{24.7±1.4} &24.8±1.2
			\\
			CenGCN &\underline{84.1±0.5} &68.1±1.1 &\underline{38.1±1.6} &24.5±1.1 &\underline{26.2±1.6}
			\\
			SAug &\textbf{86.4±0.5} &\textbf{72.1±0.8} &\textbf{40.5±1.8} &\textbf{25.4±0.8} &\textbf{28.7±1.2}
			\\
			\midrule
			GraphSAGE &81.9±1.3 &72.5±0.9 &48.0±1.6 &34.0±0.9 &30.9±1.1
			\\
			GAugSAGE &\underline{83.2±0.4} &\underline{75.7±0.7} &\underline{52.0±1.2} &\underline{37.1±1.1} &\underline{33.4±1.2}
			\\
			SAugSAGE &\textbf{91.6±1.2} &\textbf{79.1±0.6} &\textbf{56.3±1.0} &\textbf{46.7±0.8} &\textbf{36.1±0.7}
			\\
			\midrule
			GAT &76.4±1.6 &64.4±1.5 &47.5±1.3 &31.4±1.7 &26.9±1.0
			\\
			GAugGAT &\underline{77.3±0.9} &\underline{66.7±1.2} &\underline{49.7±1.6} &\underline{32.4±1.5} &\underline{27.4±1.2}
			\\
			SAugGAT &\textbf{77.7±1.0} &\textbf{67.7±1.8} &\textbf{56.6±1.5} &\textbf{33.9±1.1} &\textbf{27.6±1.1}
			\\
			\bottomrule
		\end{tabular}
	}
	\caption{Micro-F1 scores on overall node classification. In each comparison group, the best result is bolded and the second best result is underlined.}
	\label{overall_nc}
\end{table}
%\vspace{-3mm}

\subsection{Ablation Study}
Based on tail node classification, we further analyze the modules in SAug using GCN backbone. 

The bottom ﬁve lines of Table~\ref{gcn_nc} lists the performance in SAug with different modules. The results demonstrate that the denoising module improves the performance of the base model to a certain extent, and performs better than the random edge removal strategy in DropEdge. Although its performance improvement is not as good as PTDNet, its training cost is far lower than PTDNet, because the latter needs to train a parameter network for denoising. GAug conducts edge-dropping and edge-adding simultaneously. It performs better than our denoising module but still lower than SAug that selectively drops and adds edges.

In our discovering module, both threshold-based and topQ-based strategies can improve the model. On most datasets, SAug$_{thr}$ outperforms SAug$_{top}$ since the latent neighbors explored by the threshold operation are under a controllable similarity score range. Moreover, the generative module for tail nodes is effective on most datasets. 

\vspace{-0.5mm}
\begin{figure}[htbp]
	\centering
	\setlength{\abovecaptionskip}{0.cm}
	\subfigure[Discovering module]{
		\includegraphics[width=0.475\linewidth]{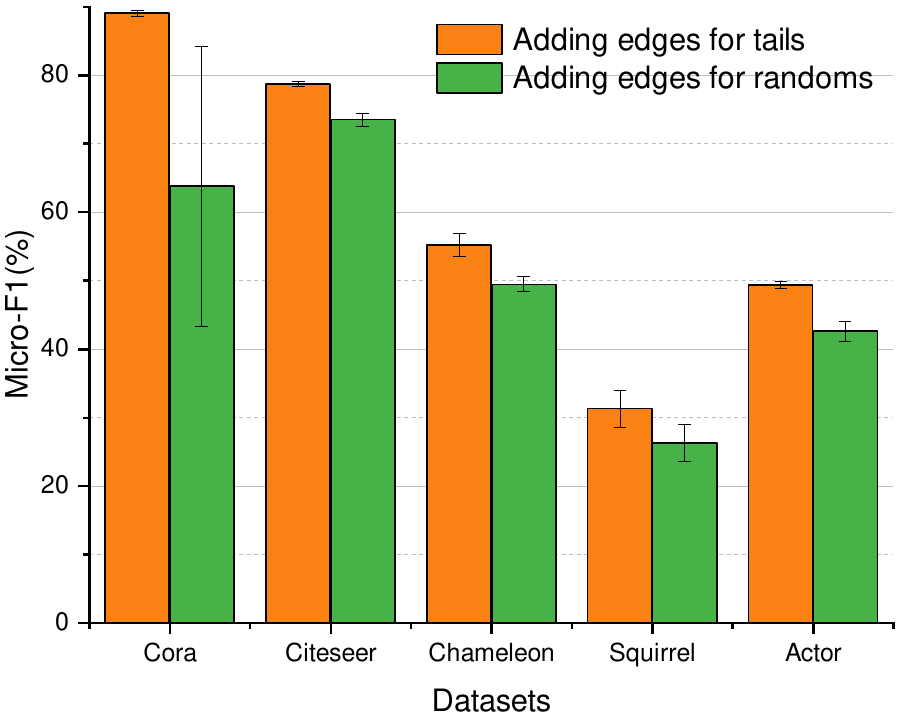}
		\label{add_edges}
	}\hspace{-0.1mm}
	\subfigure[Denoising module]
	{
		\includegraphics[width=0.475\linewidth]{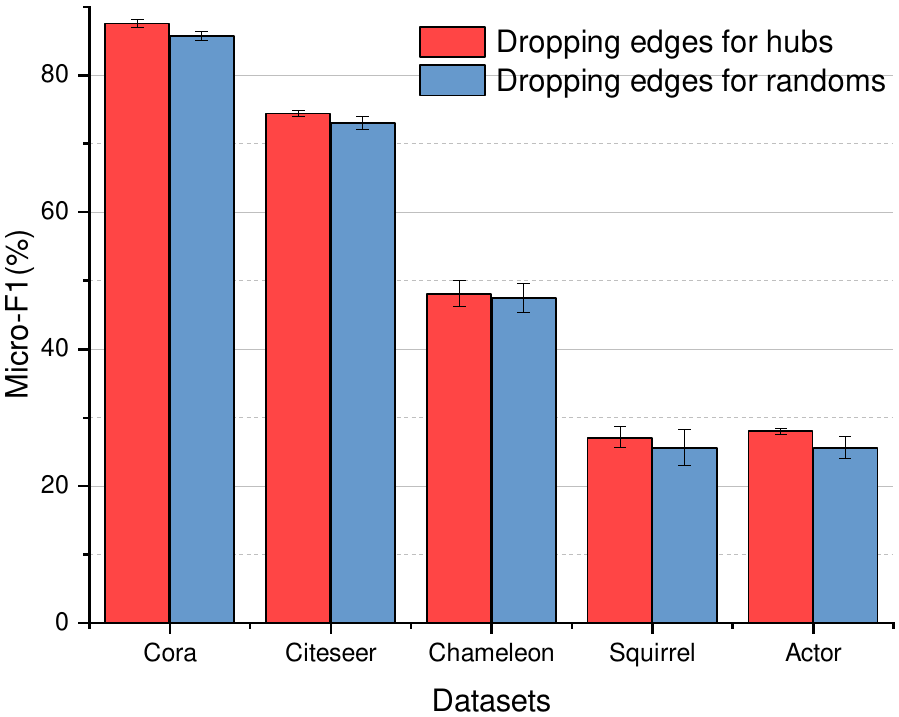}
		\label{drop_edges}
	}
	\caption{Micro-F1 comparison of the model that adds/drops edges for tails/hubs and for random nodes}
	\label{ablation}
\end{figure}
%\vspace{-2mm}

To analyze the necessity of the selective augmentation strategy, we add/drop the same number of edges for tails/hubs and random nodes, respectively. The augmentation operation is based on the threshold.  The classification results based on the augmented graph are compared in Fig~\ref{ablation}.

In the discovering module, the model adding edges for tails performs better than that adding edges for random nodes. Especially in the Cora dataset, adding edges for random nodes results in a significant degradation of model performance. The results in Fig~\ref{drop_edges} show likewise that denoising for hub nodes is superior to denoising for random nodes. 

\subsection{Hyperparameter Tuning}
\begin{figure}[htbp]
	\centering
	\setlength{\abovecaptionskip}{0.cm}
	\subfigure[Hyperparameter $P$]{
		\includegraphics[width=0.475\linewidth]{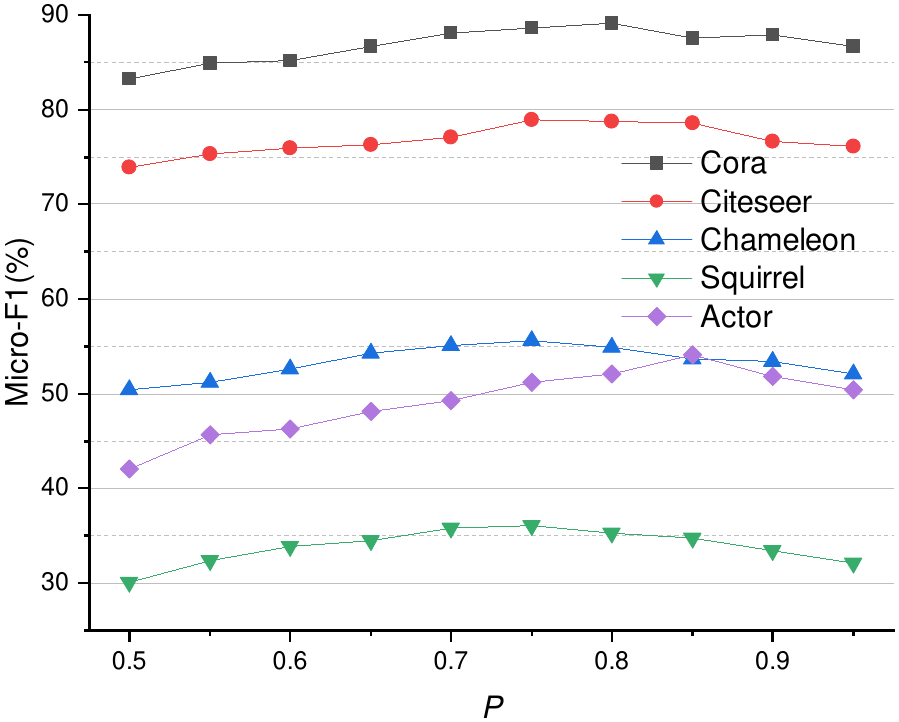}
		\label{hyper_P}
	}\hspace{-0.1mm}
	\subfigure[Hyperparameter $Q$]
	{
		\includegraphics[width=0.475\linewidth]{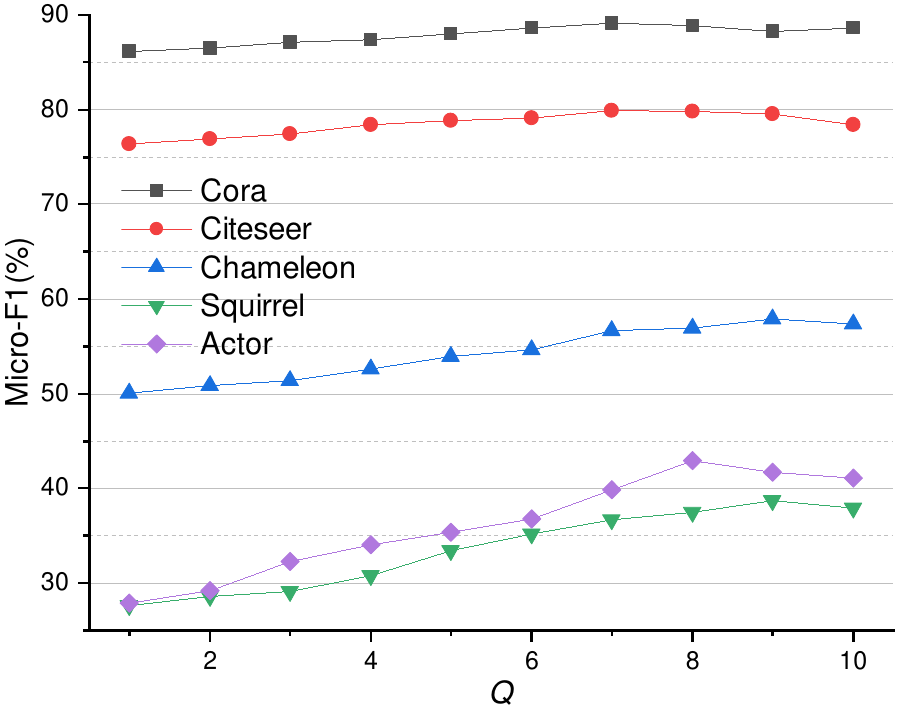}
		\label{hyper_Q}
	}
	\caption{Hyperparameters tuning in tail node classification}
	\label{hyperparameter}
\end{figure}

In this section, hyperparameters $P$ and $Q$ mentioned in the threshold operation and the topQ operation are tuned respectively to search the optimal setting. $P$ is tuned within the range of \{0.05, 0.1, ..., 0.5\} and $Q$ is tuned within the range of \{1, 2, ..., 10\}. The results are shown in Fig.~\ref{hyperparameter}.

\vspace{-0.5mm}
%\vspace{-2mm}
It demonstrates that the optimal value of $P$ and $Q$ on five datasets is almost within the range of [0.75, 0.85] and [7, 9], respectively. The model with lower $P$ or higher $Q$ may introduce noise neighbors to tail nodes while the model with higher $P$ or lower $Q$ may leads to the insufﬁcient exploration of latent neighbors.  The complete results of hyperparameter tuning are summarized in Appendix~\ref{DAug_hyper}.

\subsection{Link Prediction}
Finally, we compare all the methods on the link prediction task. The dataset is divided based on node pairs, and the ratio of training set, validation set, and testing set is 7:1:2. At each training epoch, unconnected node pairs are sampled as negative samples, with the same number of positive samples The AUC scores of TailGNN, CenGCN and SAug are shown in Table~\ref{lp_gcn}.

\begin{table}[h]
	\centering
	\resizebox{\linewidth}{!}{
		\begin{tabular}{lrrrrr}
			\toprule
			Methods &Cora &Citeseer &Chameleon &Squirrel &Actor
			\\
			\midrule
			GCN &95.6±0.4 &93.8±0.5 &95.4±0.2 &94.5±0.8 &90.1±0.3
			\\
			GCN w/ $Gen$ &97.1±0.3 &97.7±0.4 &96.0±0.1 &95.6±0.5 &90.6±0.3
			\\
			\midrule
			TailGCN &96.7±0.3 &\underline{96.9±0.4} &95.9±0.2 &\underline{95.4±0.7} &\underline{90.7±0.2}
			\\
			CenGCN &\underline{97.3±0.4} &95.3±0.2 &\underline{96.1±0.4} &95.2±0.3 &90.3±0.4
			\\
			\midrule
			%Denoise &95.5±0.3 &93.9±0.5 &95.6±0.3 &95.0±0.5 &90.2±0.4
			SAug$_{thr}$ w/o $Gen$ &96.2±0.3 &95.1±0.3 &96.2±0.1 &94.7±0.6 &89.7±0.3
			\\
			SAug$_{thr}$ &96.9±0.2 &97.4±0.3 &\textbf{96.7±0.2} &95.7±0.6 &90.3±0.2
			\\
			SAug$_{top}$ w/o $Gen$ &97.3±0.2 &97.4±0.3 &96.4±0.1 &94.7±0.6 &90.6±0.3
			\\
			SAug$_{top}$ &\textbf{97.7±0.4} &\textbf{98.3±0.3} &96.6±0.3 &\textbf{95.7±0.4} &\textbf{90.8±0.2}
			\\
			\midrule
			\midrule
			GraphSAGE &94.6±0.3 &95.4±0.5 &95.1±0.2 &94.4±0.1 &88.1±0.4
			\\
			GraphSAGE w/ $Gen$ &95.5±0.3 &96.8±0.5 &96.0±0.1 &94.3±0.2 &88.2±0.5
			\\
			\midrule
			TailSAGE &\underline{96.1±0.4} &\underline{96.4±0.3} &\underline{96.2±0.5} &\underline{94.7±0.2} &\underline{89.2±0.3}
			\\
			\midrule
			%Denoise &94.6±0.6 &95.5±0.5 &96.1±0.3 &94.6±0.1 &88.1±0.3
			SAug$_{thr}$ w/o $Gen$ &94.6±0.5 &96.0±0.6 &94.4±0.5 &93.3±0.3 &88.4±0.4
			\\
			SAug$_{thr}$ &95.1±0.4 &96.9±0.5 &96.1±0.3 &93.6±0.3 &87.8±0.4
			\\
			SAug$_{top}$ w/o $Gen$ &96.0±0.2 &97.1±0.3 &96.5±0.2 &\textbf{95.1±0.4} &\textbf{89.9±0.3}
			\\
			SAug$_{top}$ &\textbf{96.3±0.4} &\textbf{97.3±0.2} &\textbf{96.7±0.2} &94.3±0.2 &89.3±0.4
			\\
			\midrule
			\midrule
			GAT &93.0±0.5 &92.6±0.7 &94.1±0.3 &92.4±0.3 &84.7±0.4
			\\
			GAT w/ $Gen$ &94.5±0.5 &96.0±0.4 &94.3±0.3 &93.3±0.5 &85.0±0.5
			\\
			\midrule
			TailSAGE &\underline{94.4±0.2} &\underline{95.8±0.4} &\underline{94.2±0.3} &\underline{93.1±0.5} &\underline{85.2±0.6}
			\\
			\midrule
			%Denoise &92.9±0.3 &92.7±0.7 &94.2±0.5 &92.6±0.4 &84.2±0.6
			SAug$_{thr}$ w/o $Gen$ &92.7±0.7 &91.6±0.6 &92.1±0.6 &92.7±0.3 &84.1±0.6
			\\
			SAug$_{thr}$ &94.5±0.3 &95.3±0.4 &92.6±0.6 &\textbf{93.4±0.5} &\textbf{85.7±0.4}
			\\
			SAug$_{top}$ w/o $Gen$ &93.8±0.2 &94.3±0.4 &94.2±0.2 &92.3±0.5 &83.7±0.8
			\\
			SAug$_{top}$ &\textbf{94.8±0.5} &\textbf{96.3±0.4} &\textbf{94.5±0.2} &93.1±0.4 &84.7±0.5
			\\
			\bottomrule
		\end{tabular}
	}
	\caption {AUC scores on link prediction. The bolded data in each column is the best result, and the underlined data are the best results in comparative models..}
	\label{lp_gcn}
\end{table}
%\vspace{-2mm}
In link prediction, SAug also achieves the competitive performance. Note that SAug$_{top}$ overall performs better than SAug$_{thr}$. As the training of the link prediction requires unconnected node pairs as the negative samples, adding too much neighbors may affect the negative sampling. The number of edges added by the topQ operation is usually less than that by the threshold operation. Therefore, the topQ operation is more suitable for the link prediction task. Besides, the generative module is of great help in improving model’s performance in predicting links. It is probably because the generated pseudo nodes are also counted in negative sampling, increases the diversity of negative samples.
%\vspace{-1mm}
\section{Conclusion}
In this paper, we propose a novel graph augmentation method SAug to solve the problem of the structural imbalance on the graph. We first define hub and tail nodes based on the Pagerank value and accordingly propose a Pagerank-based sampling strategy to identify the two types of nodes. Subsequently, we propose a selective structural augment strategy to alleviate the structural imbalance between hubs and tails and improve their representation. Additionally, we generate pseudo neighbors for tails to further enrich their neighborhood. SAug is veriﬁed to achieve superior performance against other baseline models via extensive experiments.

\appendix
\section{Reproducibility\label{reproducibility}}
\subsection{Dataset details\label{datasets_details}}
The statistics of all datasets are summarized in Table~\ref{datasets}, and the Pagerank values in Table~\ref{datasets} are the results after multiplying by $10^3$ for the convenience of presentation.
\paragraph{Citation networks.} Cora~\cite{mccallum2000automating} and Citeseer~\cite{giles1998citeseer} are citation network datasets whose nodes are papers published in the field of computer science, features are bag-of-word vectors of the corresponding paper title and edges are citation relationship between two papers. Following~\cite{GCN}, we treat the citation links as (undirected) edges and construct a binary and symmetric adjacency matrix A.
\paragraph{Wikipedia networks.} Chameleon and Squirrel~\cite{rozemberczki2021multi} are two page-to-page networks on specific topics in Wikipedia pages. In these datasets, nodes repersent web pages and edges represent mutual links between two pages. Actor~\cite{tang2009social} is the actor-only induced subgraph of the film-director-actor-writer network, each node corresponds to an actor, and the edge between two nodes denotes co-occurrence on the same Wikipedia page. These datasets are preprocessed following~\cite{pei2020geom}.

\begin{table*}[htbp]
	\centering
	\begin{tabular}{lrrrrrrr} 
		\toprule
		Datasets     &Nodes    &Edges    &Features  &Labels &Max\_Pagerank &Min\_Pagerank &Avg\_Pagerank
		\\ 
		\midrule
		Cora   	  &\num{2708}   &\num{5278}   &\num{1433}    &7 &11.36 &0.0591 &0.3692\\
		CiteSeer  &\num{3327}   &\num{4552}   &\num{3703}   &6 &5.045 &0.0524 &0.3005\\
		% Pubmed    &19717  &\num{44324}    &500   &3 &1.559 &0.0087 &0.0507\\
		Squirrel  &\num{5201}   &\num{217073}    &\num{2089}   &5 &5.682 &0.0295 &0.1922\\
		Chameleon &\num{2277}   &\num{36101}   &\num{2325}   &5 &19.38 &0.0682 &0.4391\\
		Actor     &\num{7600}   &\num{30019}    &932   &5 &21.14 &0.0251 &0.1315\\
		\bottomrule
	\end{tabular}
	\caption{Statistics of datasets.}  
	\label{datasets}  
\end{table*}

\subsection{Implementation details\label{implementation}}
PyTorch~\cite{paszke2019pytorch} and PyG~\cite{fey2019fast} is used to implement SAug. Codes of GCN, GraphSAGE and GAT in the pretraining step and experiments are implemented referringto PyG implementation of GCN, GraphSAGE and GAT, respectively. All experiments are all conducted on a Linux server with a GeForce RTX3090-24GB. The softwares we use for experiments include Python 3.8.8, Pytorch 1.12.1, Pytorch-cluster 1.6.0, Pytorch-Scatter 2.0.9, Pytorch-Sparse 0.6.15, Pytorch-Geometric 2.1.0 and CUDA 11.6.1. 

\section{Hyperparameter Tuning}
\subsection{Base GNN models}
All GNN models are implemented in PyG with the Adam optimizer. Based on the recommended setting in original papers, we further tune the basic hyperparameters to obtain optimal performance. In the node classification task, for all base GNN models, we adopt a three-layer architecture with a hidden dimension of 32 and a output of the number of classes. In the link prediction task, we adopt a two-layer achitecture with a hidden layer of 32 and a output dimension of 16. Additionally, for GAT, we use three attention heads in each layer, and apply a dropout rate of 0.5 for features and attention; for GraphSAGE, we adopt a mean-pooling over the neighbors during aggregation. All GNN models have the learning rate of 0.01 and the weight decay rate of 0.0005. Moreover, for all tasks, we set the coefficients of regularization as 0.0001.

\begin{figure}[htbp]
	\centering
	\subfigure[$K$ in SAugGCN on tail node classification]{
		\includegraphics[width=0.45\linewidth]{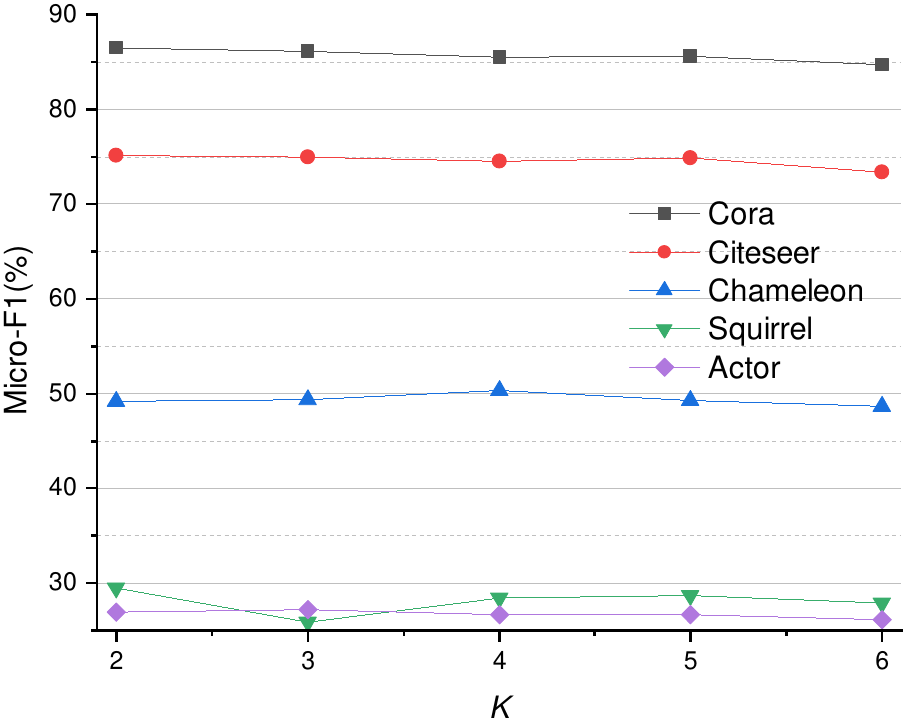}
		\label{nc_K_gcn}
	}\hspace{-0.1mm}
	\subfigure[$K$ in SAugGCN on link prediction]{
		\includegraphics[width=0.45\linewidth]{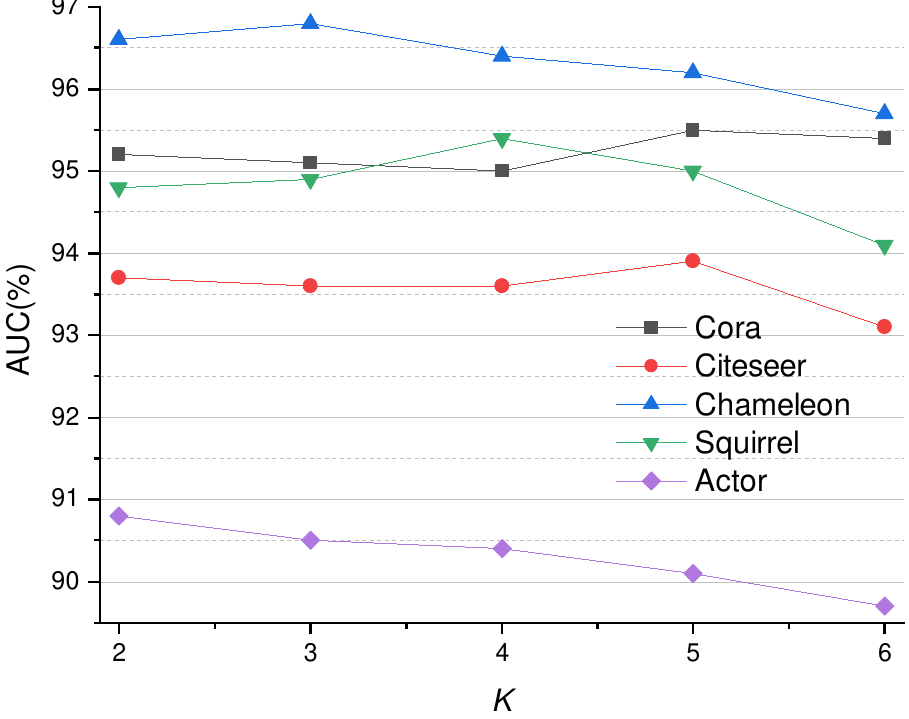}
		\label{lp_K_gcn}
	}
	\\	
	\subfigure[$K$ in SAugSAGE on tail node classification]
	{
		\includegraphics[width=0.45\linewidth]{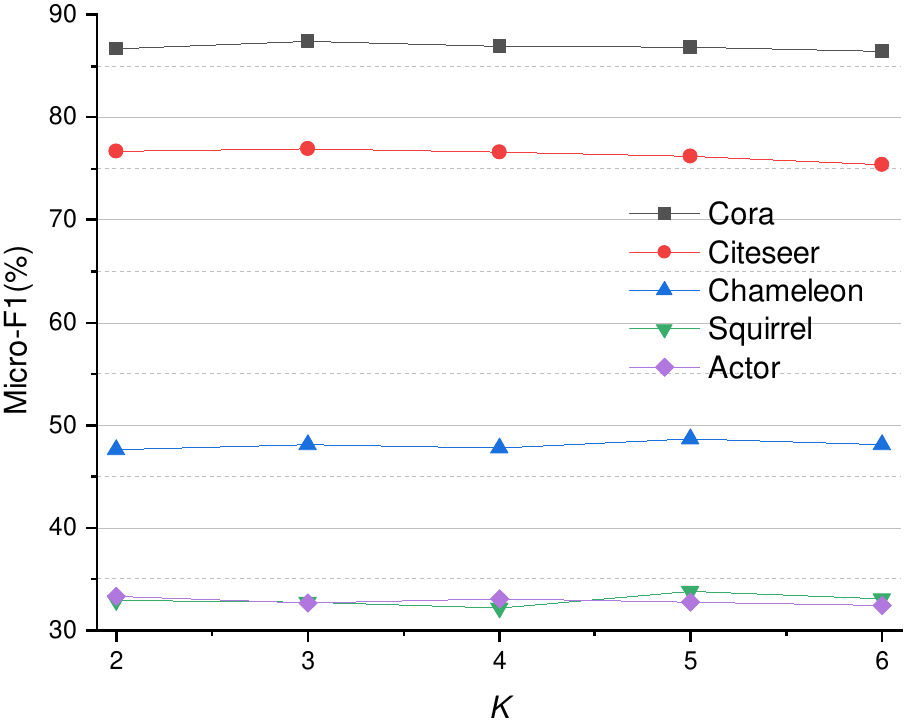}
		\label{nc_K_sage}
	}
	\subfigure[$K$ in SAugSAGE on link prediction]
	{
		\includegraphics[width=0.45\linewidth]{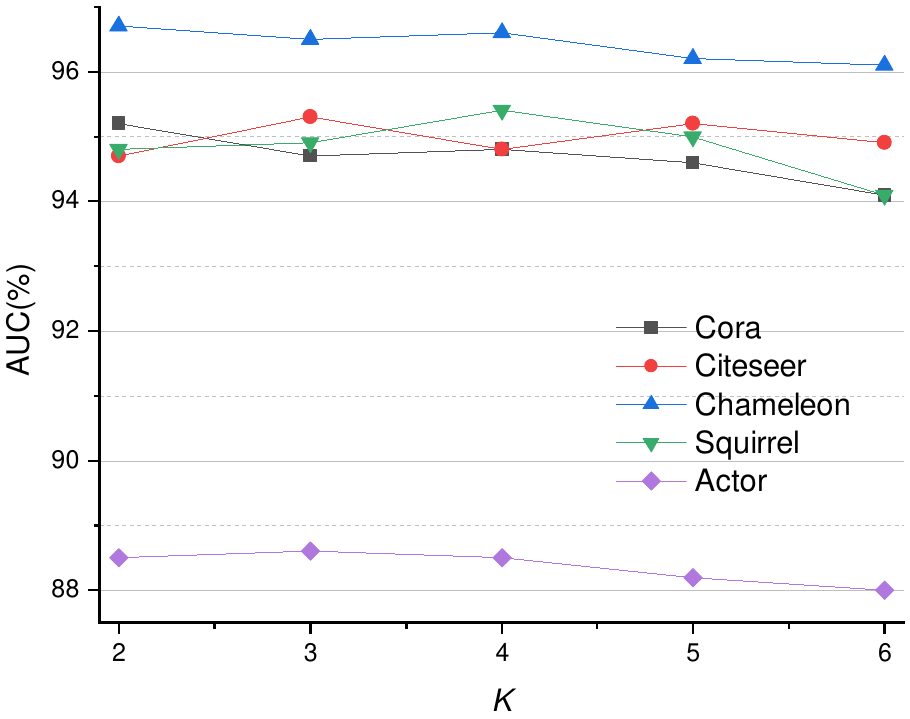}
		\label{lp_K_sage}
	}
	\\
	\subfigure[$K$ in SAugGAT on tail node classification]
	{
		\includegraphics[width=0.45\linewidth]{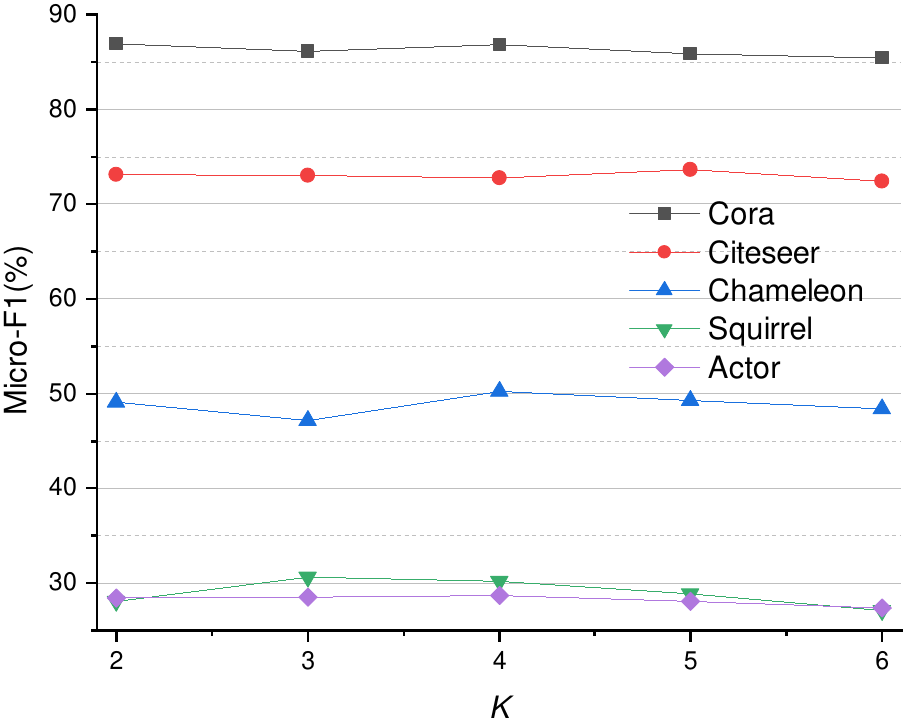}
		\label{nc_K_gat}
	}
	\subfigure[$K$ in SAugGAT on link prediction]
	{
		\includegraphics[width=0.45\linewidth]{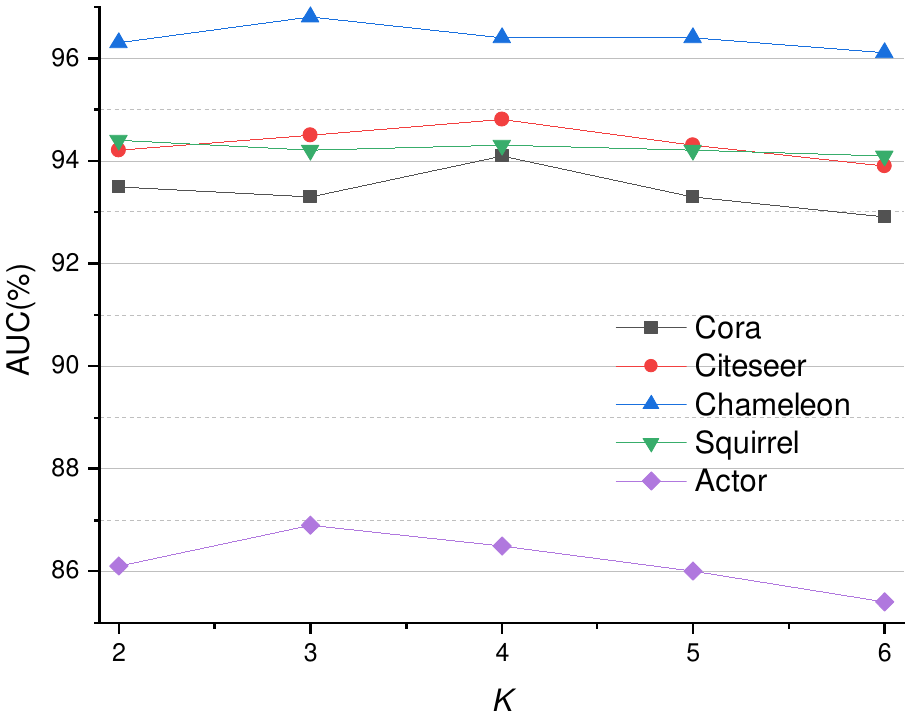}
		\label{lp_K_gat}
	}
	\caption{Tuning of Hyparameter $K$}
	\label{hub_K}
\end{figure}

\begin{figure}[h]
	\centering
	\subfigure[$M$ in SAugGCN on tail node classification]{
		\includegraphics[width=0.45\linewidth]{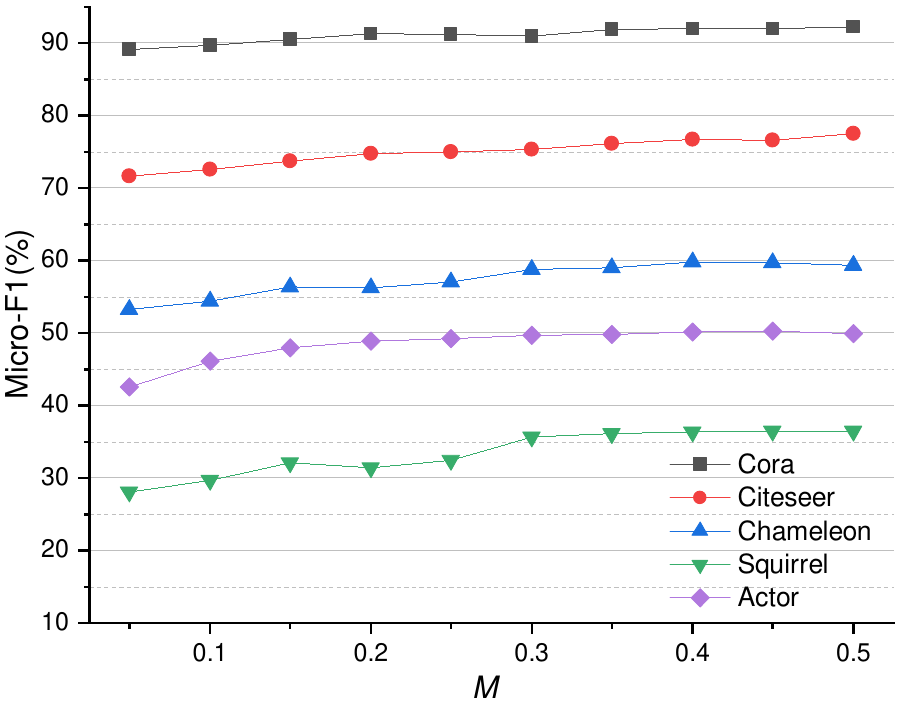}
		\label{tail_ratio_gcn}
	}\hspace{1mm}
	\subfigure[$M$ in SAugGCN on link prediction]{
		\includegraphics[width=0.45\linewidth]{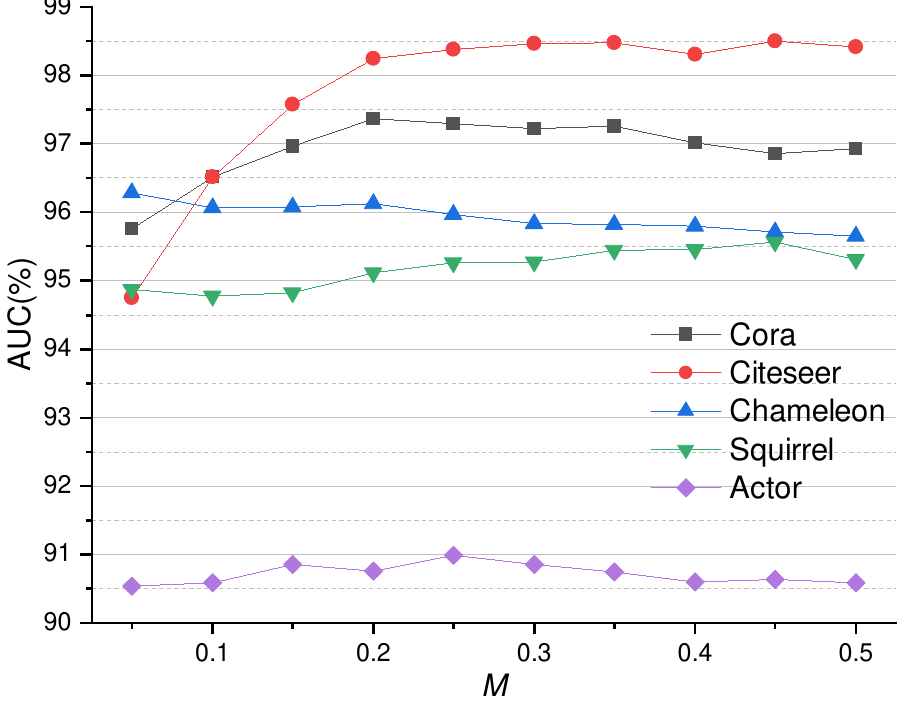}
		\label{lp_tail_ratio_gcn}
	}
	\\
	\subfigure[$M$ in SAugSAGE on tail node classification]
	{
		\includegraphics[width=0.45\linewidth]{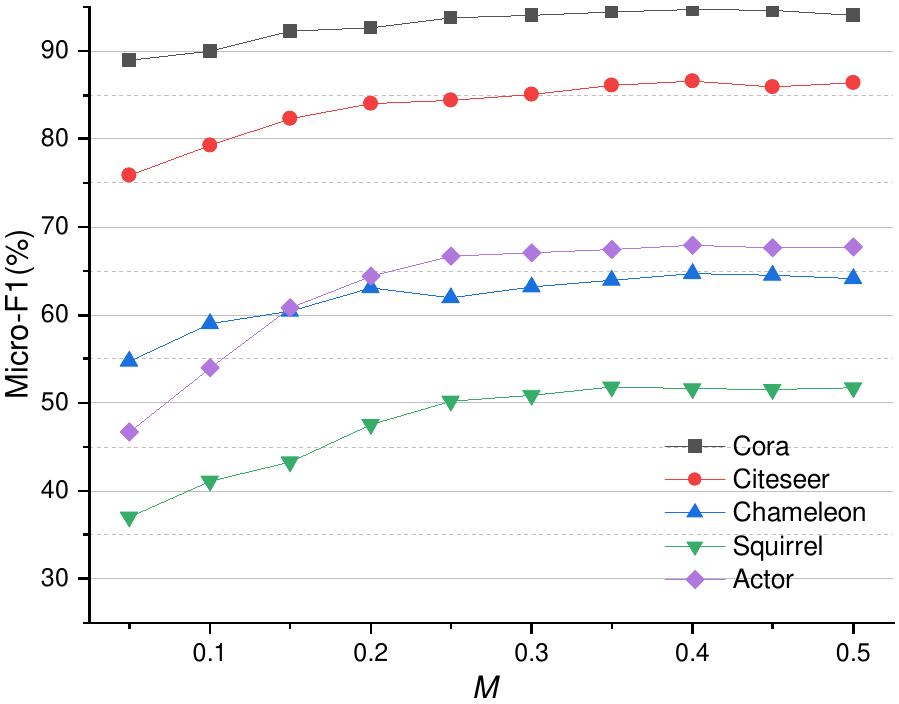}
		\label{tail_ratio_sage}
	}\hspace{1mm}
	\subfigure[$M$ in SAugSAGE on link prediction]
	{
		\includegraphics[width=0.45\linewidth]{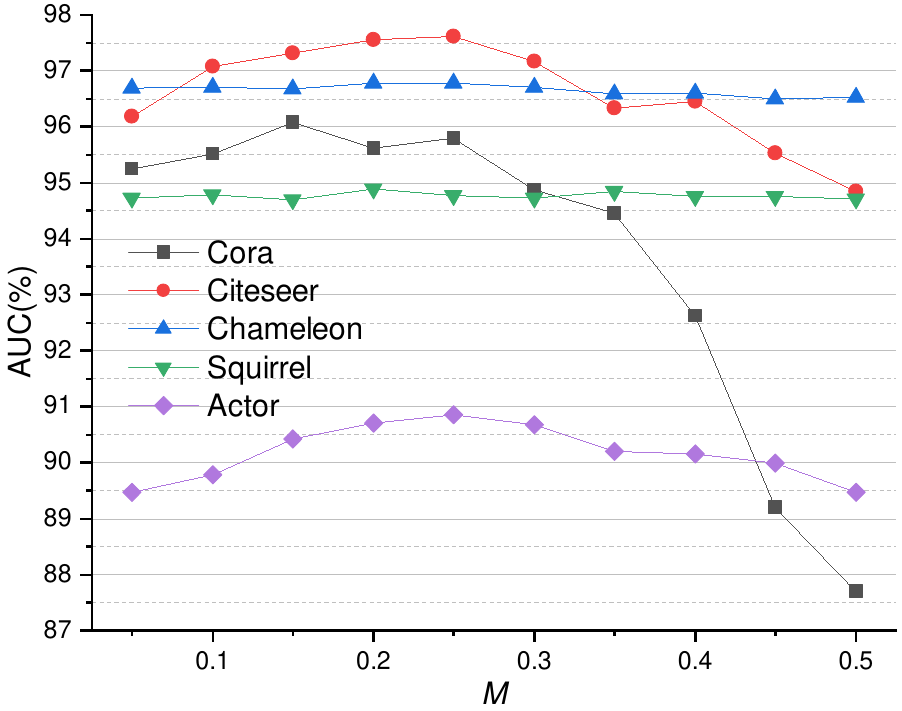}
		\label{lp_tail_ratio_sage}
	}
	\\
	\subfigure[$M$ in SAugGAT on tail node classification]
	{
		\includegraphics[width=0.45\linewidth]{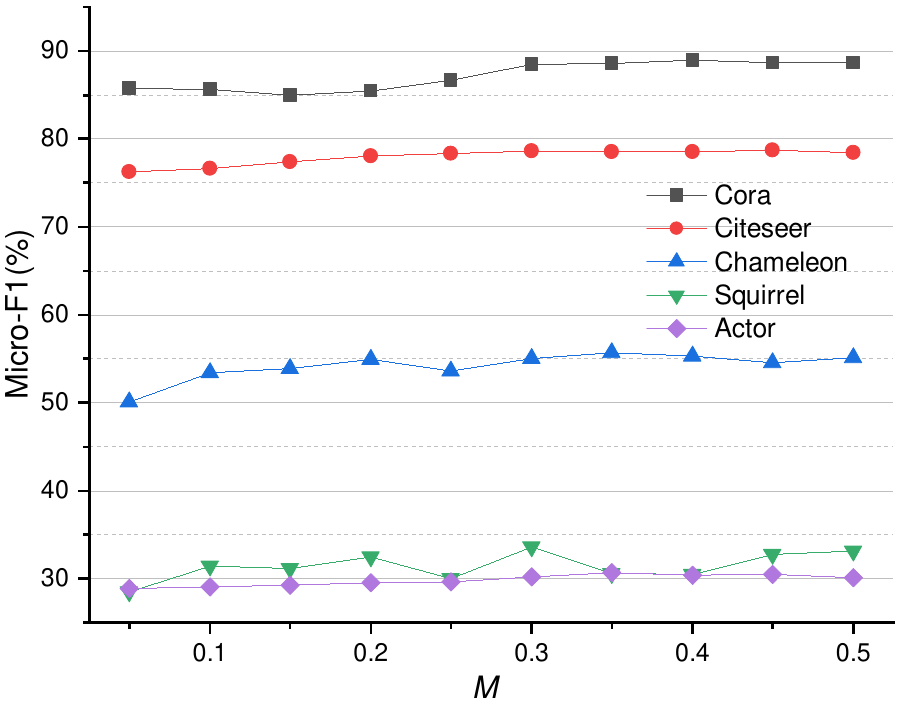}
		\label{tail_ratio_gat}
	}\hspace{1mm}
	\subfigure[$M$ in SAugGAT on link prediction]
	{
		\includegraphics[width=0.45\linewidth]{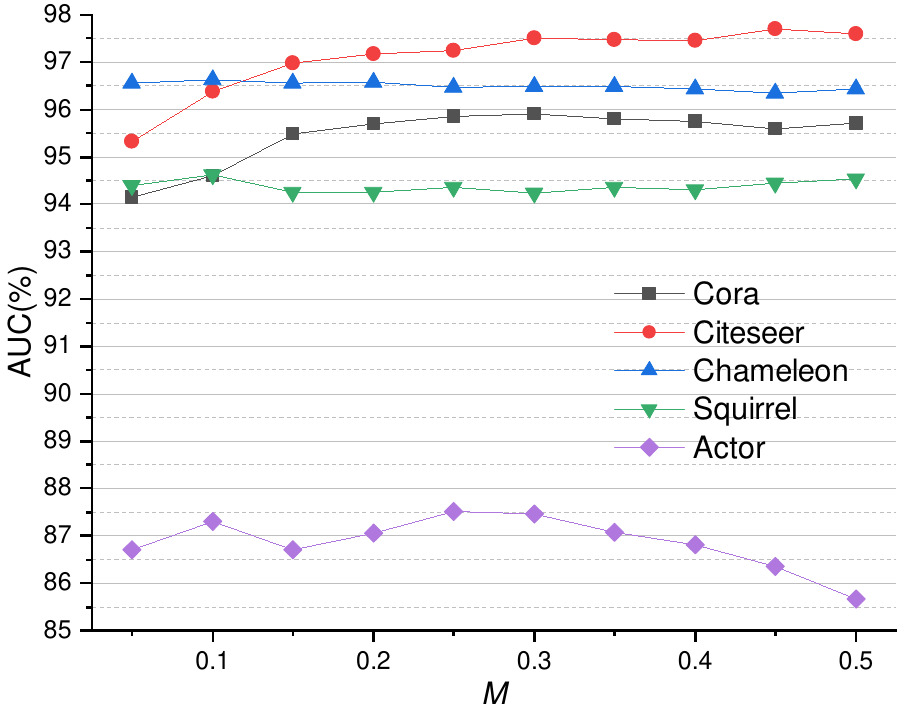}
		\label{lp_tail_ratio_gat}
	}
	\caption{Tuning of Hyparameter $M$}
	\label{tail_ratio}
\end{figure}

\begin{figure}[H]
	\centering
	\subfigure[$P$ in SAugSAGE]
	{
		\includegraphics[width=0.45\linewidth]{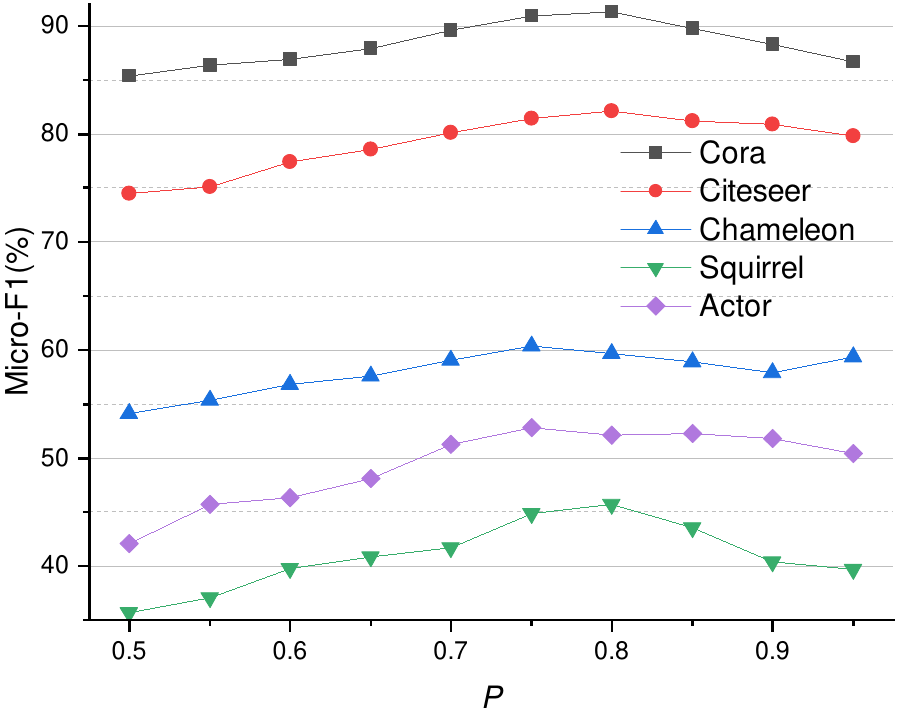}
		\label{nc_thres_sage}
	}\hspace{1mm}
	\subfigure[$Q$ in SAugSAGE]
	{
		\includegraphics[width=0.45\linewidth]{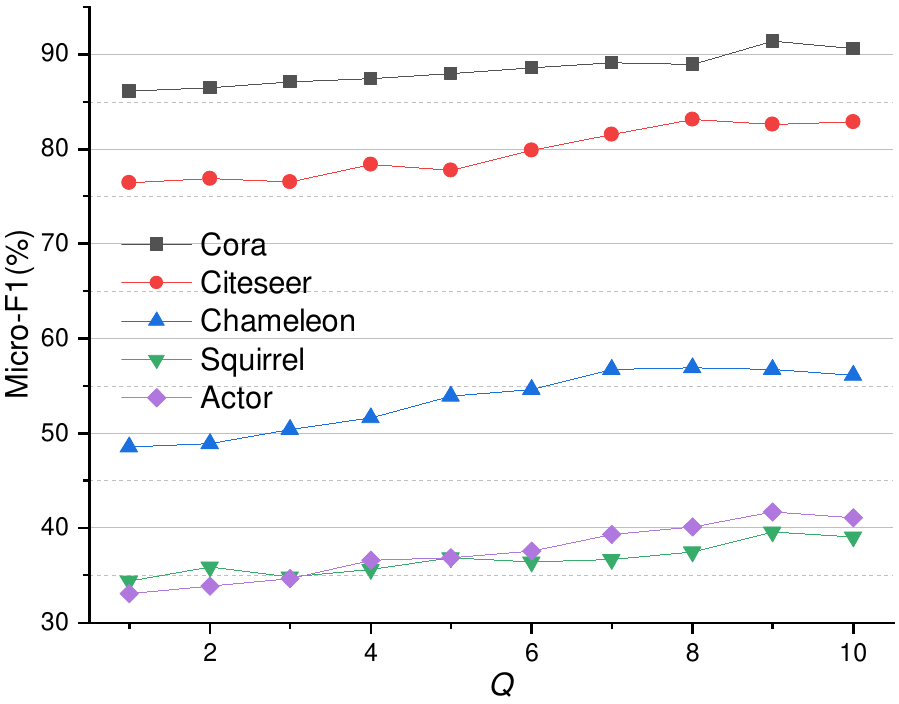}
		\label{nc_topq_sage}
	}
	\\
	\subfigure[$P$ in SAugGAT]
	{
		\includegraphics[width=0.45\linewidth]{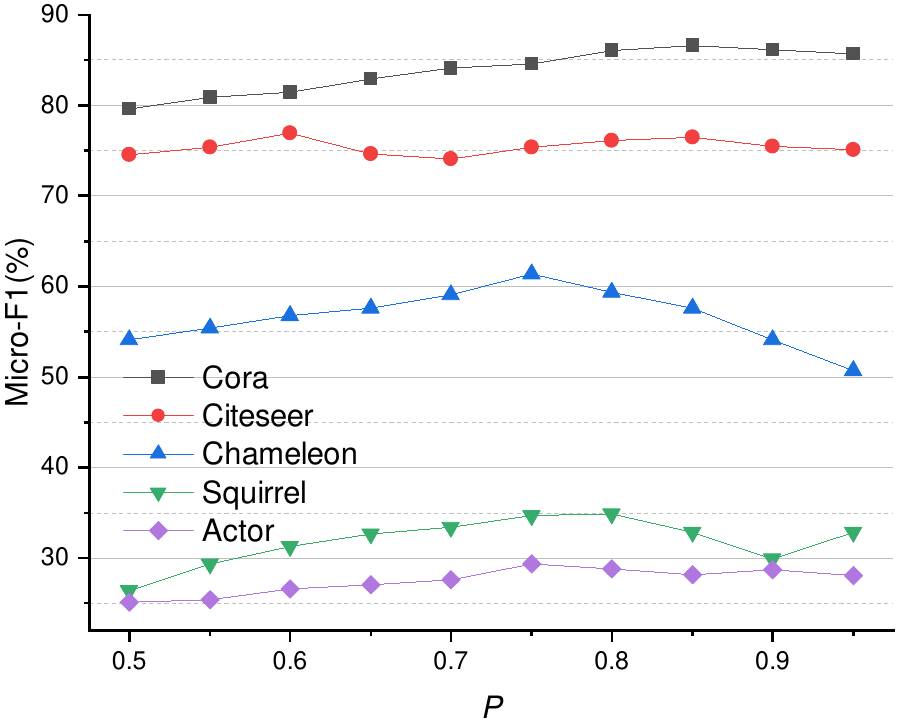}
		\label{nc_thres_gat}
	}\hspace{1mm}
	\subfigure[$Q$ in SAugGAT]
	{
		\includegraphics[width=0.45\linewidth]{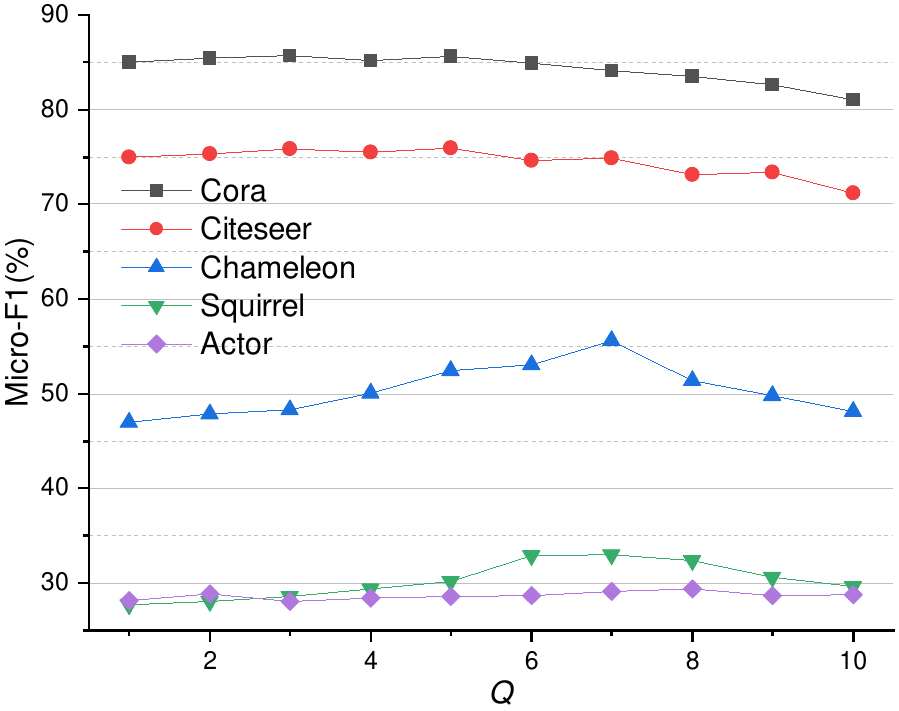}
		\label{nc_topq_gat}
	}
	\caption{Tuning of $P$ and $Q$ on tail node classification}
	\label{nc_thres}
\end{figure}

\subsection{SAug\label{DAug_hyper}}
\paragraph{Hyperparameter $K$ and $M$.}
As mentioned in Section~\ref{sample}, we have two hyperparameters $K$ and $M$ to respectively control the number of hub nodes and tail nodes, where $K$ is denoted as: $PR(v_i)\ge K*PR_{avg}$, $v_i \in \mathcal{V}_{hub}$, and $M$ is the tail node ratio. $K$ is tuned within the range of \{2, 3, 4, 5, 6\}, $M$ is tuned within the range of \{0.05, 0.1,..., 0.5\}. The tuning results of $K$ and $M$ on node classification and link prediciton are shown in Figure~\ref{hub_K} and Figure~\ref{tail_ratio}, respectively.

\begin{figure}[h]
	\centering
	\subfigure[$P$ in SAugGCN]
	{
		\includegraphics[width=0.45\linewidth]{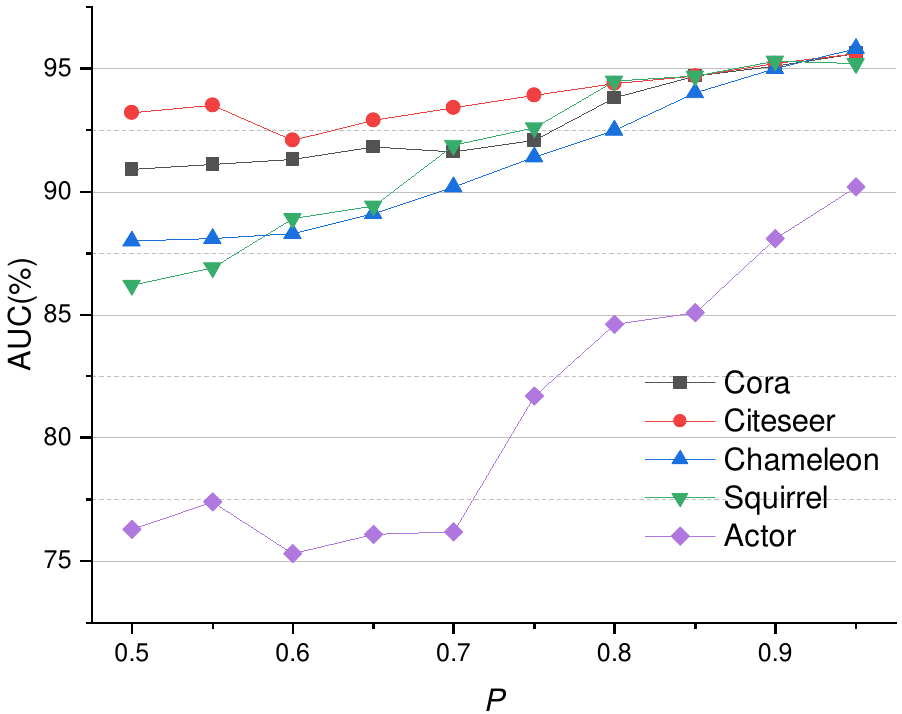}
		\label{lp_thres_gcn}
	}
	\subfigure[$Q$ in SAugGCN]
	{
		\includegraphics[width=0.45\linewidth]{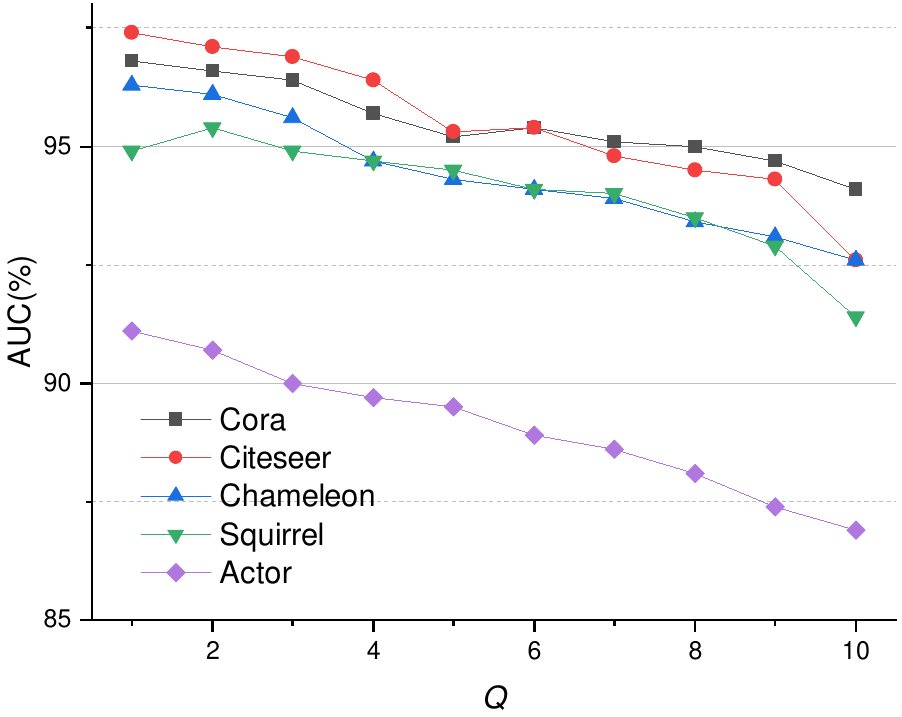}
		\label{lp_topq_gcn}
	}
	\\
	\subfigure[$P$ in SAugSAGE]
	{
		\includegraphics[width=0.45\linewidth]{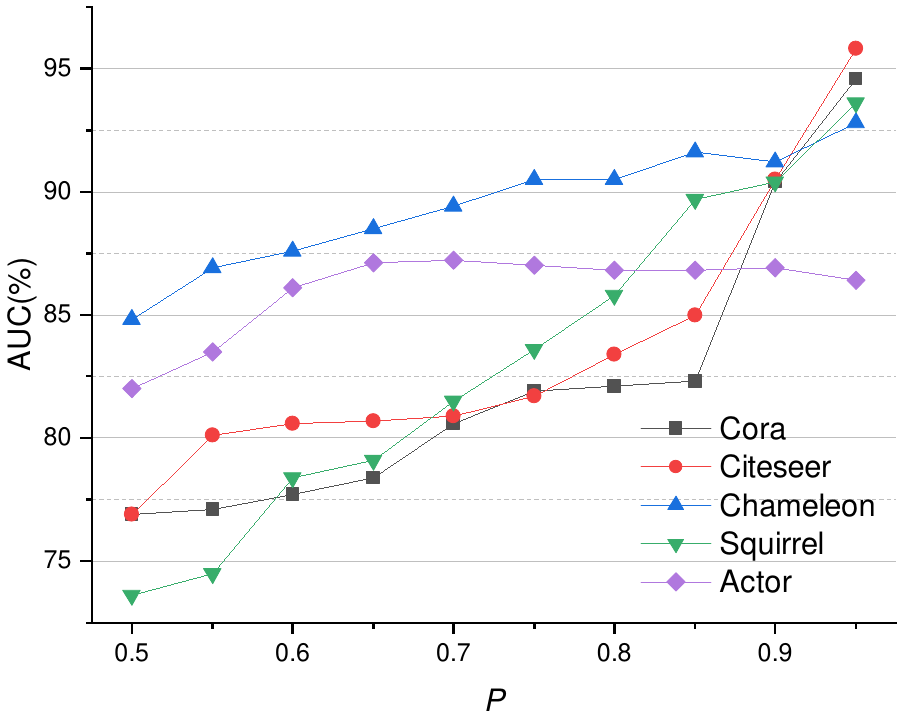}
		\label{lp_thres_sage}
	}
	\subfigure[$Q$ in SAugSAGE]
	{
		\includegraphics[width=0.45\linewidth]{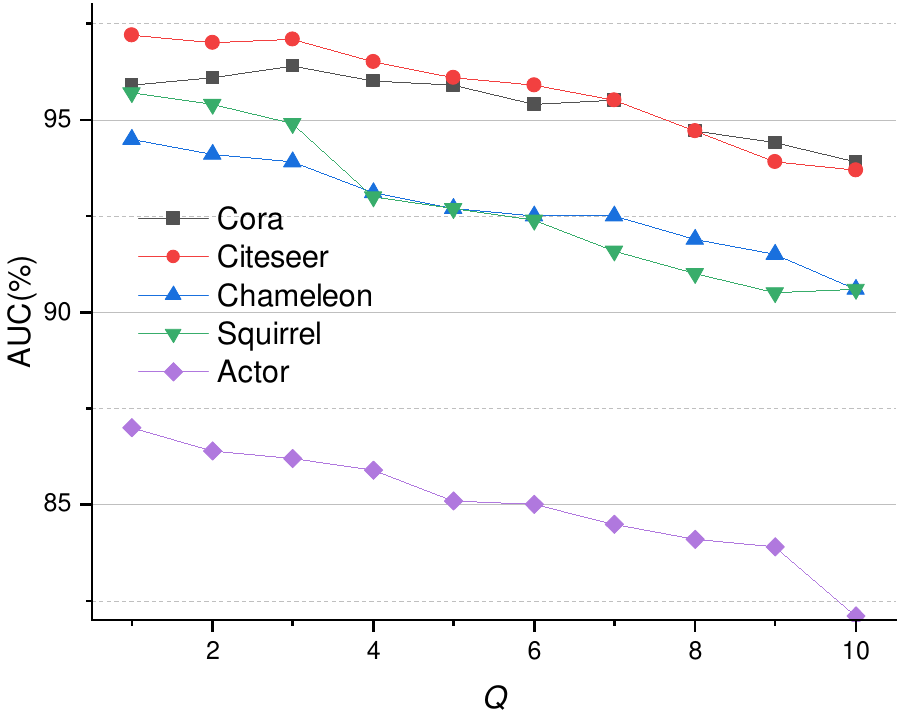}
		\label{lp_topq_sage}
	}
	\\
	\subfigure[$P$ in SAugGAT]
	{
		\includegraphics[width=0.45\linewidth]{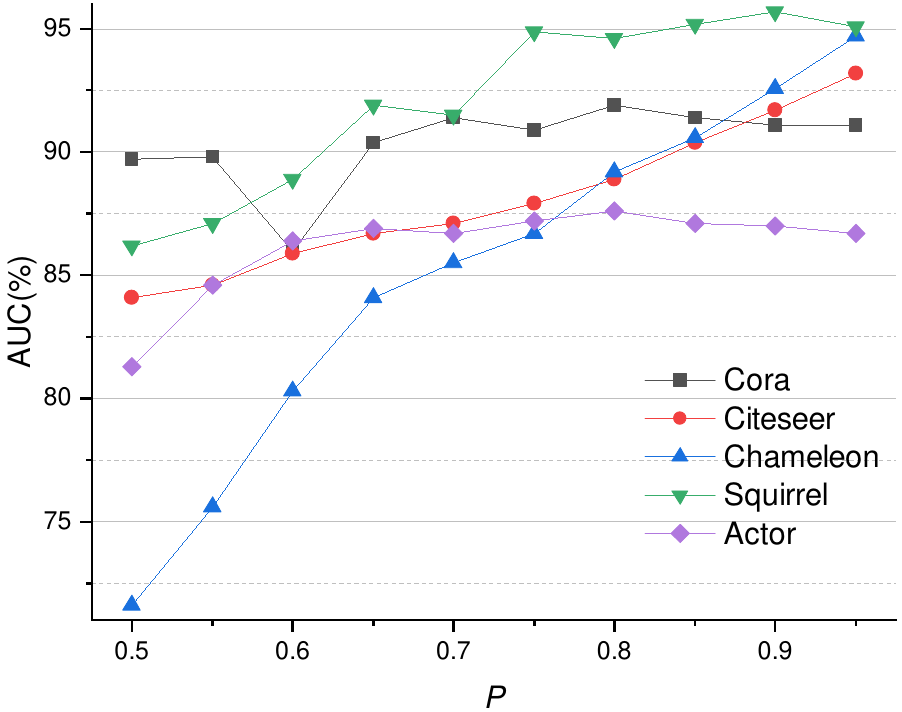}
		\label{lp_thres_gat}
	}
	\subfigure[$Q$ in GAugGAT]
	{
		\includegraphics[width=0.45\linewidth]{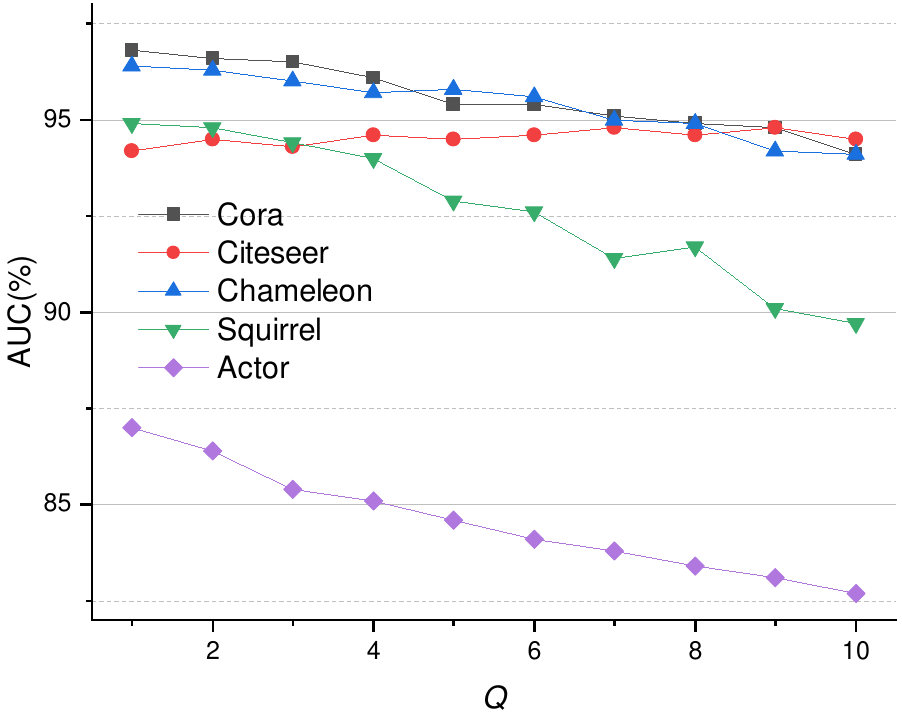}
		\label{lp_topq_gat}
	}
	\caption{Tuning of $P$ and $Q$ on link prediction}
	\label{lp_thres}
\end{figure}

\begin{figure}[h]
	\centering
	\subfigure[$L$ in SAugGCN on tail node classification]{
		\includegraphics[width=0.45\linewidth]{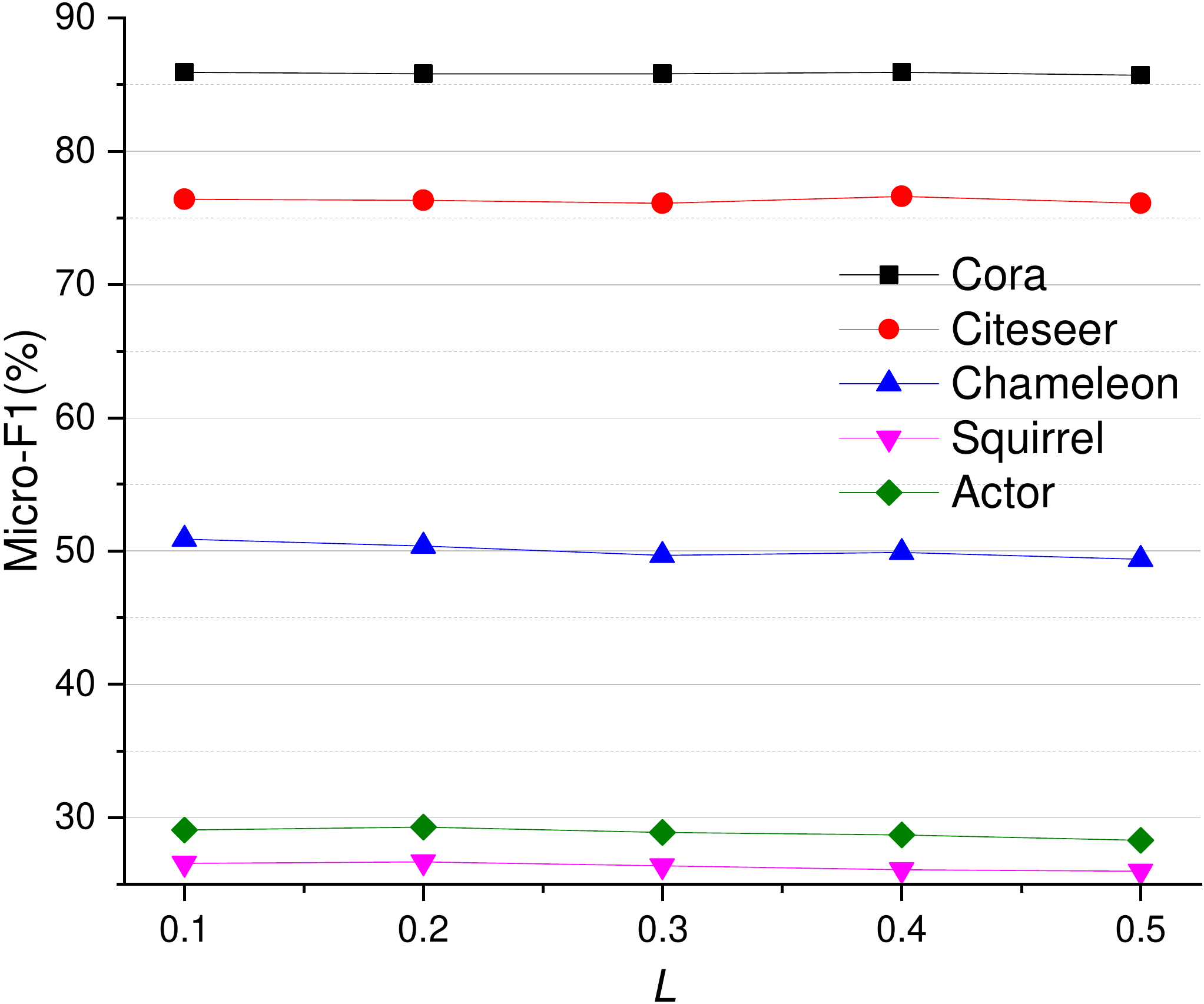}
		\label{nc_denoise_gcn}
	}\hspace{-0.1mm}
	\subfigure[$L$ in SAugGCN on link prediction]{
		\includegraphics[width=0.45\linewidth]{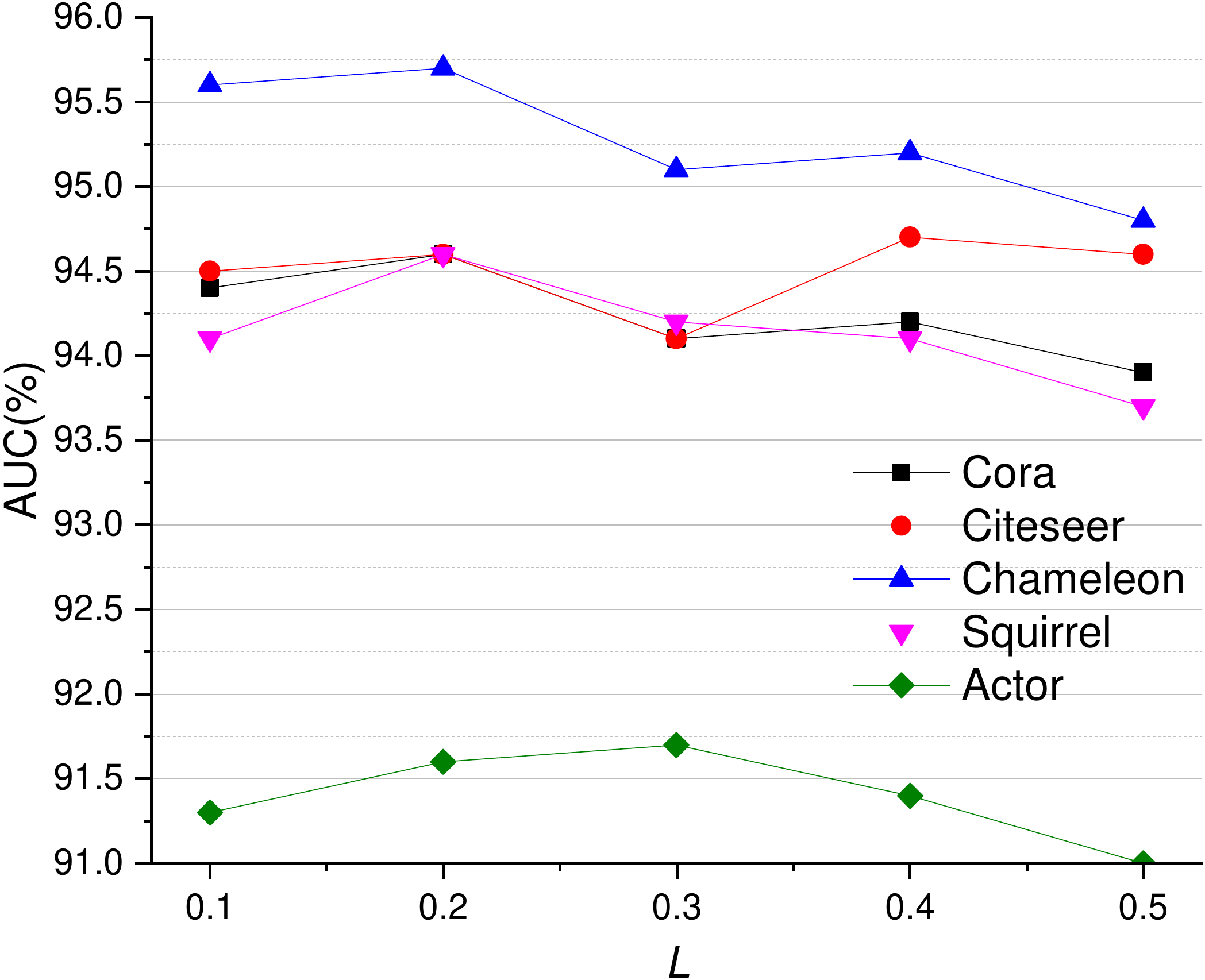}
		\label{lp_denoise_gcn}
	}\hspace{-0.1mm}
	\\	
	\subfigure[$L$ in SAugSAGE on tail node classification]
	{
		\includegraphics[width=0.45\linewidth]{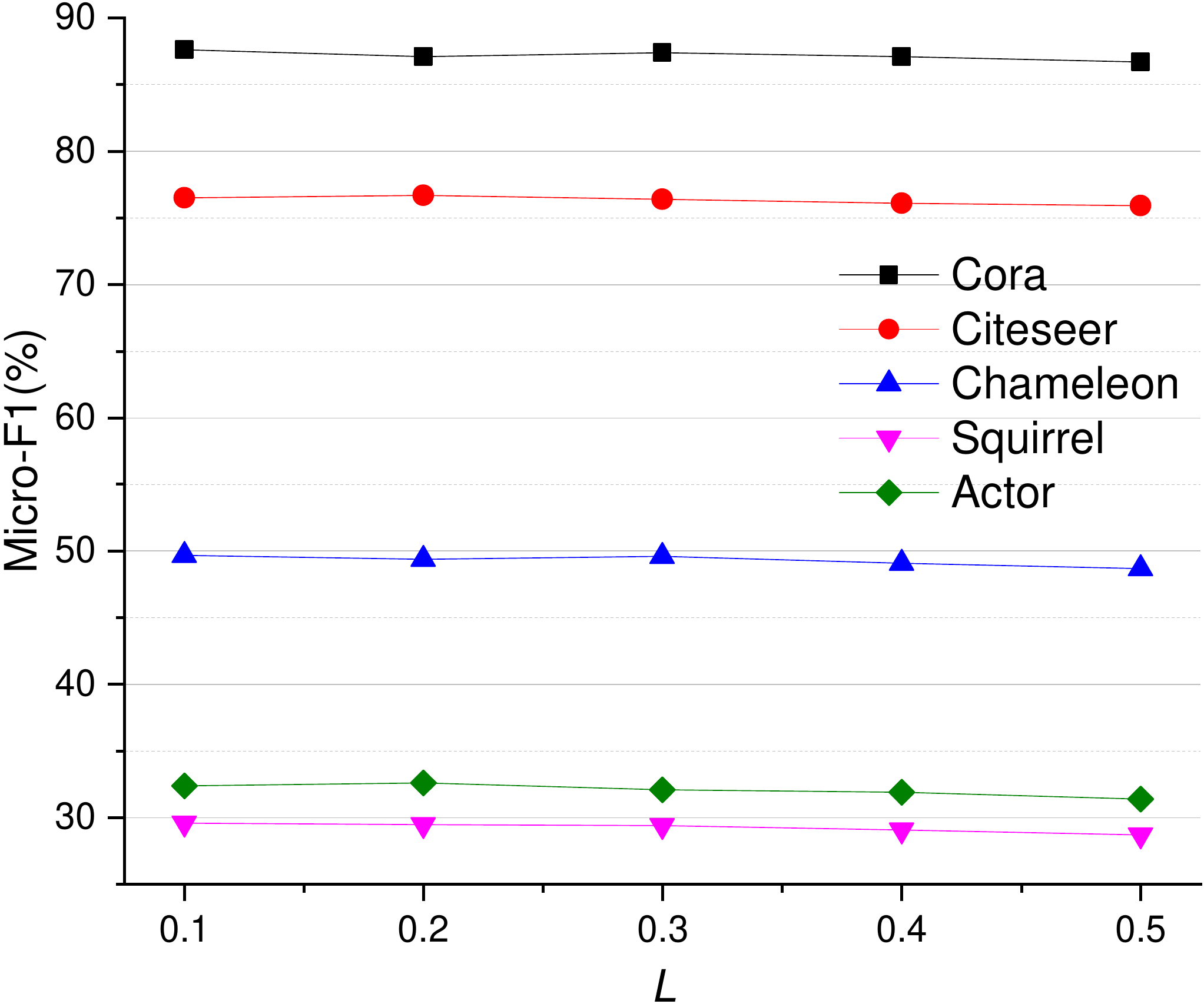}
		\label{nc_denoise_sage}
	}
	\subfigure[$L$ in SAugSAGE on link prediction]
	{
		\includegraphics[width=0.45\linewidth]{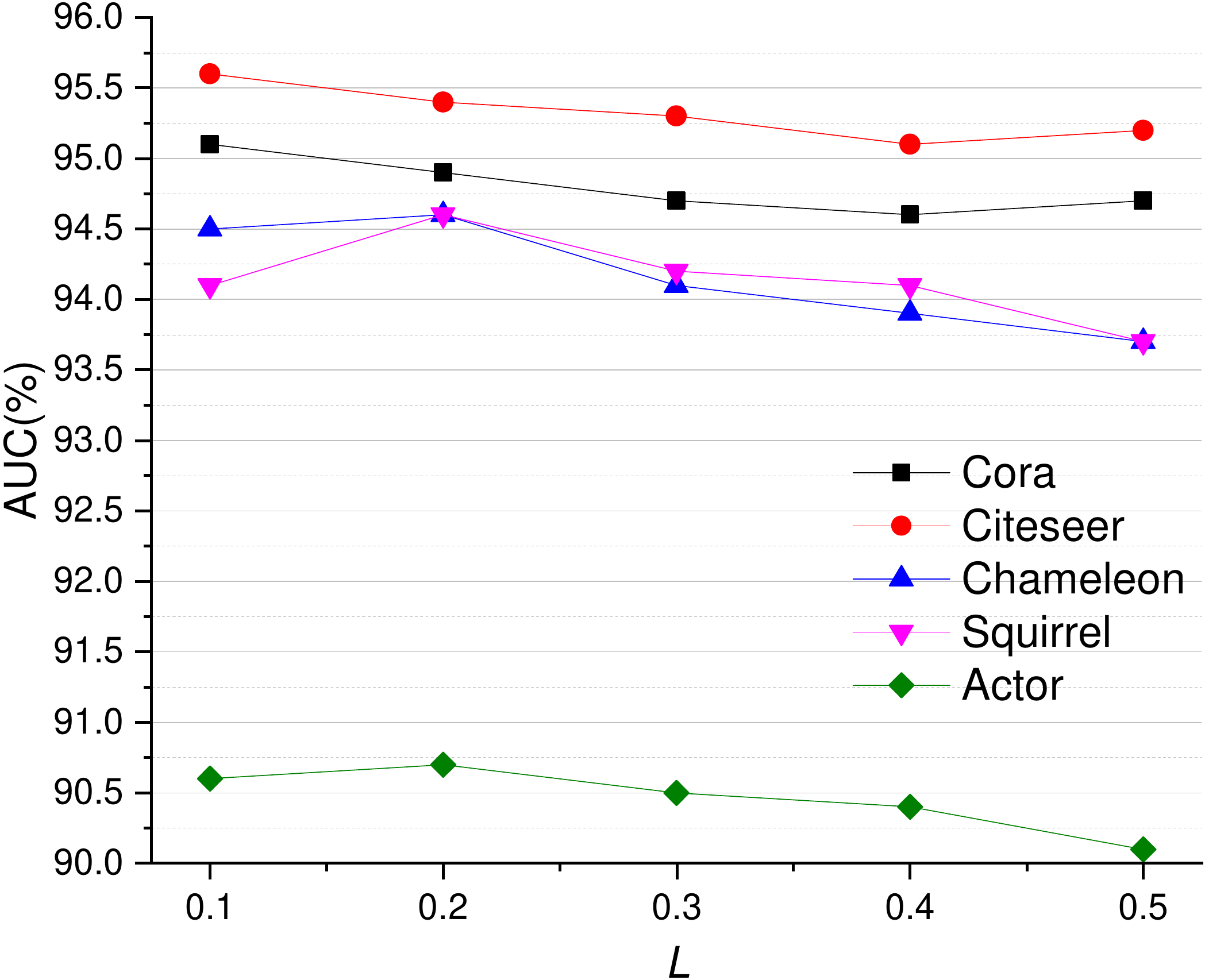}
		\label{lp_denoise_sage}
	}
	\\
	\subfigure[$L$ in SAugGAT on tail node classification]
	{
		\includegraphics[width=0.45\linewidth]{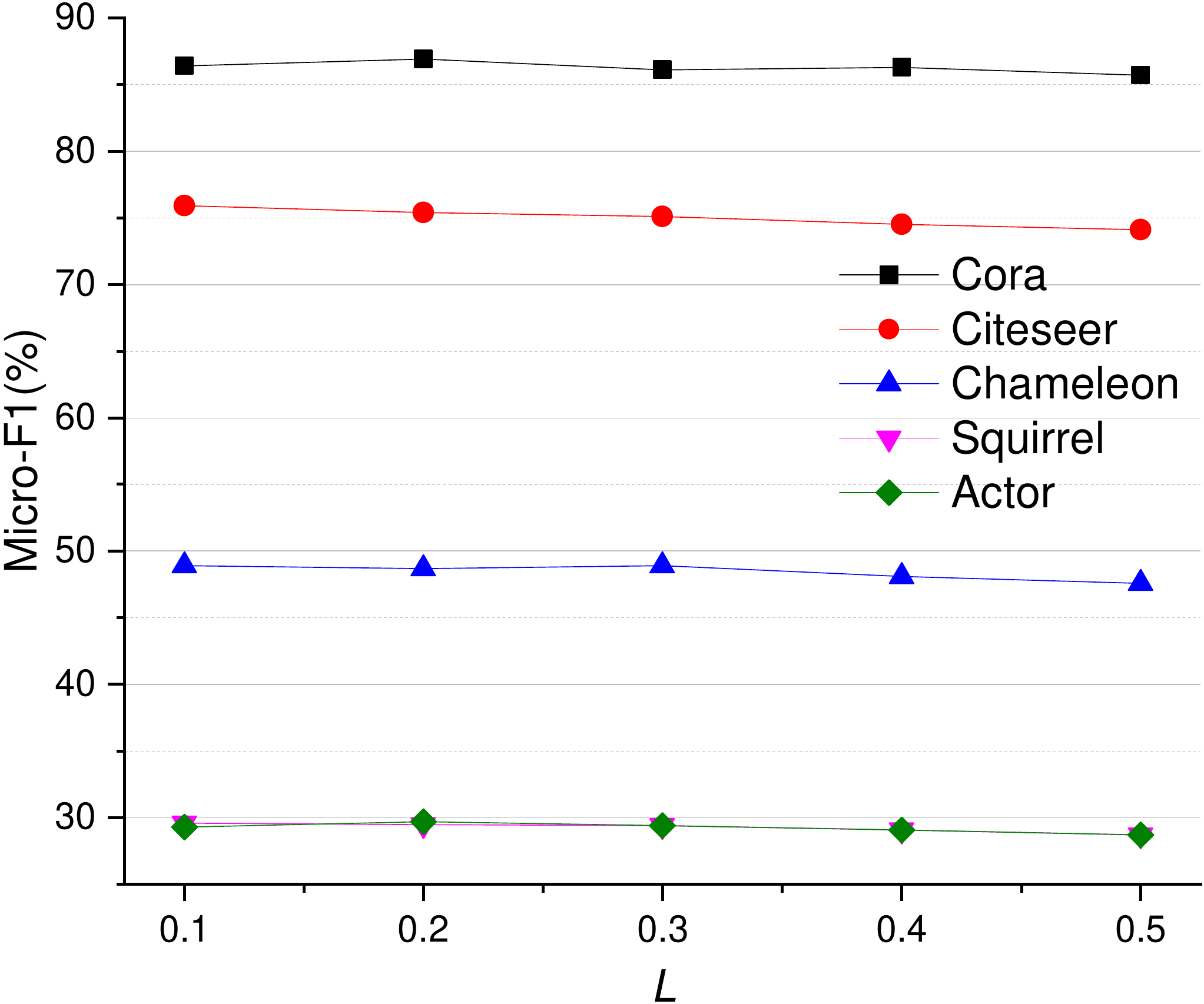}
		\label{nc_denoise_gat}
	}
	\subfigure[$L$ in SAugGAT on link prediction]
	{
		\includegraphics[width=0.45\linewidth]{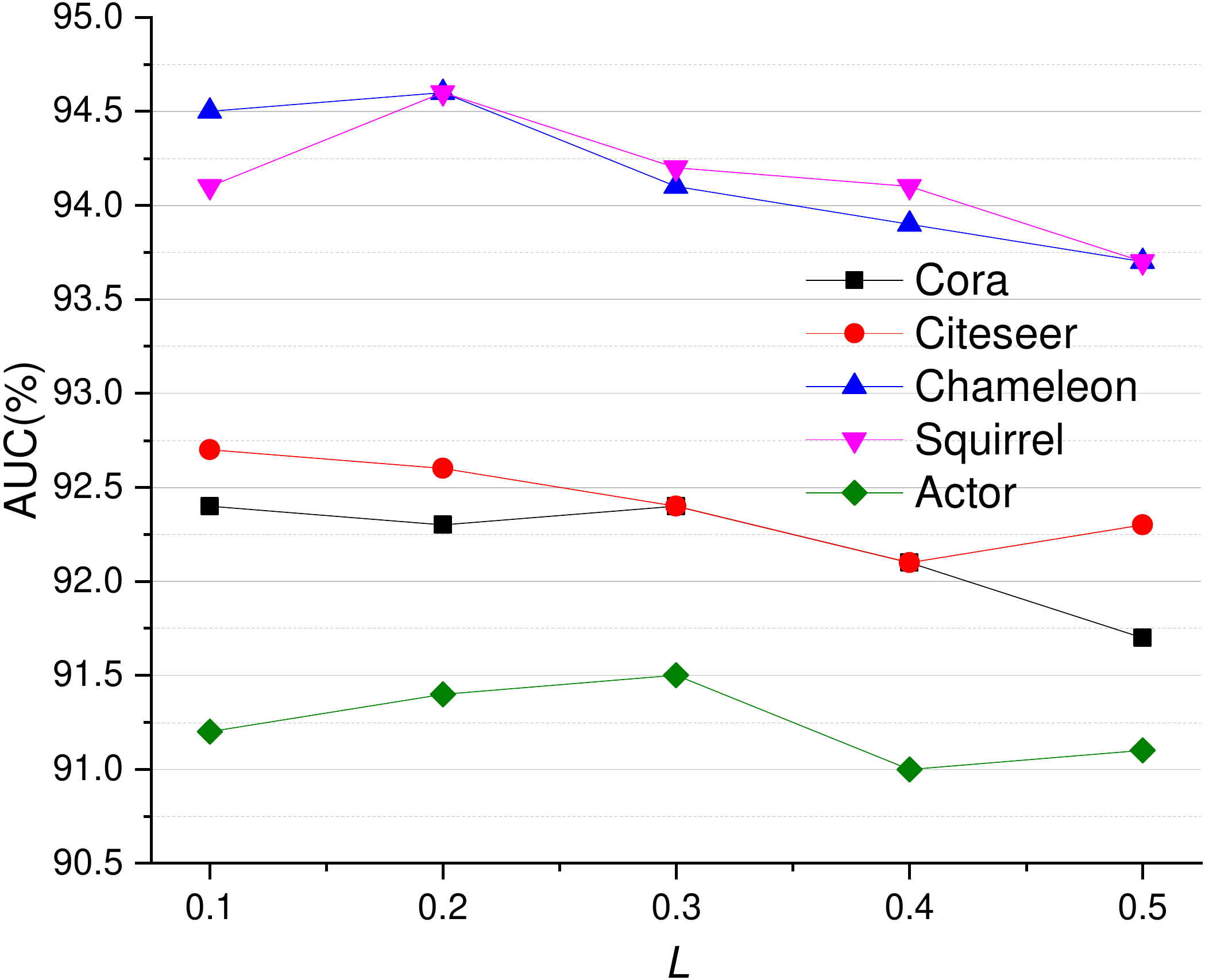}
		\label{lp_denoise_gat}
	}
	\caption{Tuning of Hyparameter $L$}
	\label{nc_denoise}
\end{figure}

\paragraph{Hyperparameter $L$, $P$ and $Q$.} 
As mentioned in Section~\ref{denoise}, $L$ is used to decide noise edges in hubs' neighbors to be dropped, $P$ and $Q$ respectively represent the threshold operation and the topQ opreation which are used to explore latent neighbors for tails. $L$ is tuned within the range of \{0.1, 0.2, 0.3, 0.4, 0.5\}, $P$ is tuned within the range \{0.05, 0.1,...,0.5\}, $Q$ is tuned within the range of \{1, 2,..., 10\}. Tunning results are shown in Figure~\ref{nc_thres}, Figure~\ref{lp_thres} and Figure~\ref{nc_denoise}, respectively.

\begin{figure}[H]
	\centering
	\subfigure[Discovering module in SAugSAGE]
	{
		\includegraphics[width=0.45\linewidth]{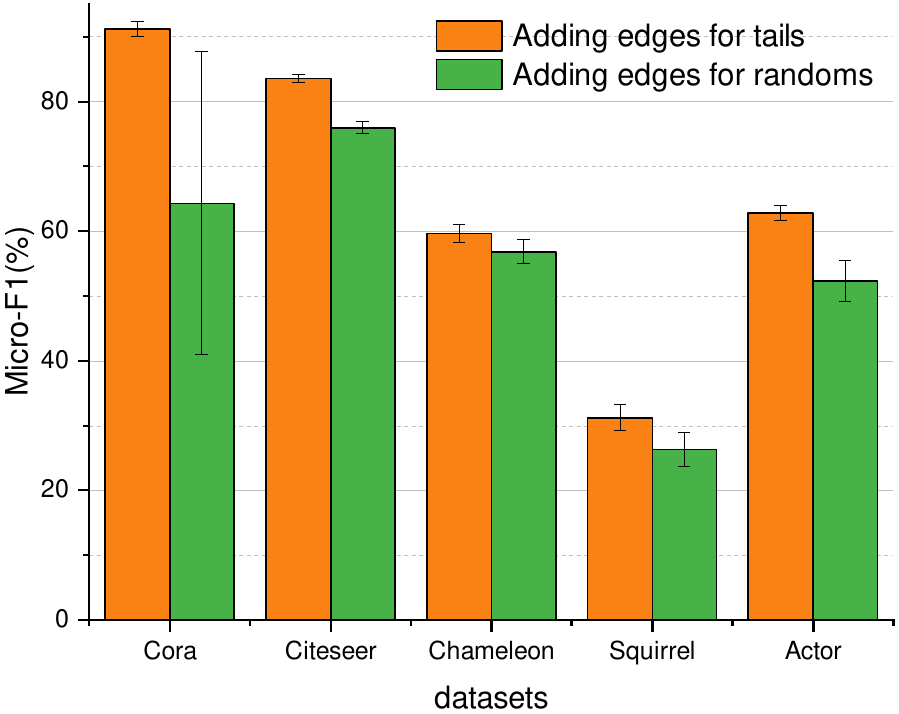}
		\label{adding_sage}
	}
	\subfigure[Denoising module in SAugSAGE]
	{
		\includegraphics[width=0.45\linewidth]{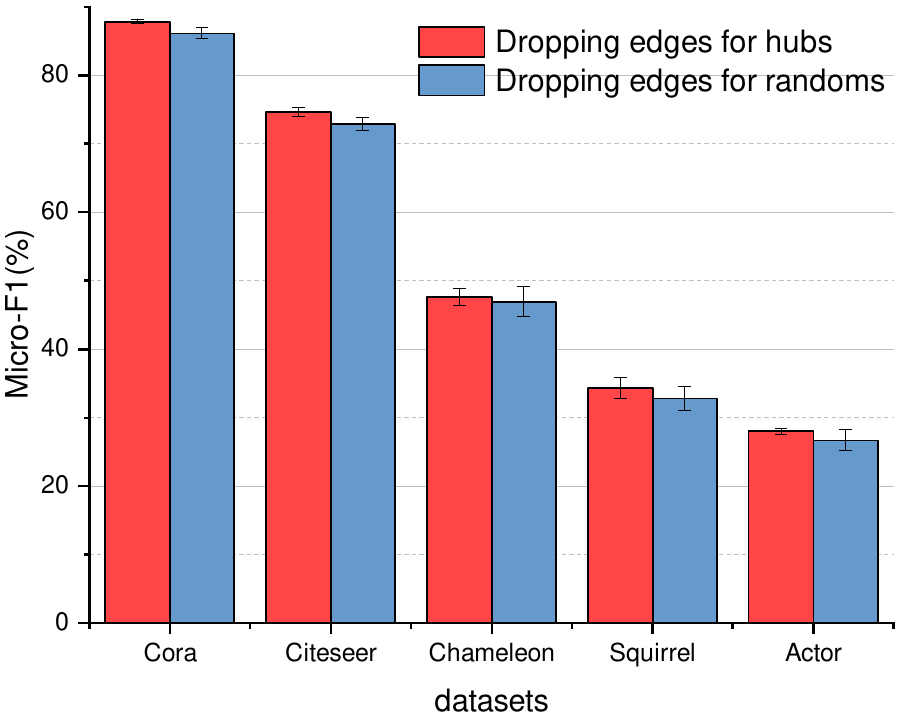}
		\label{denoise_sage}
	}
	\\
	\subfigure[Discovering module in SAugGAT]
	{
		\includegraphics[width=0.45\linewidth]{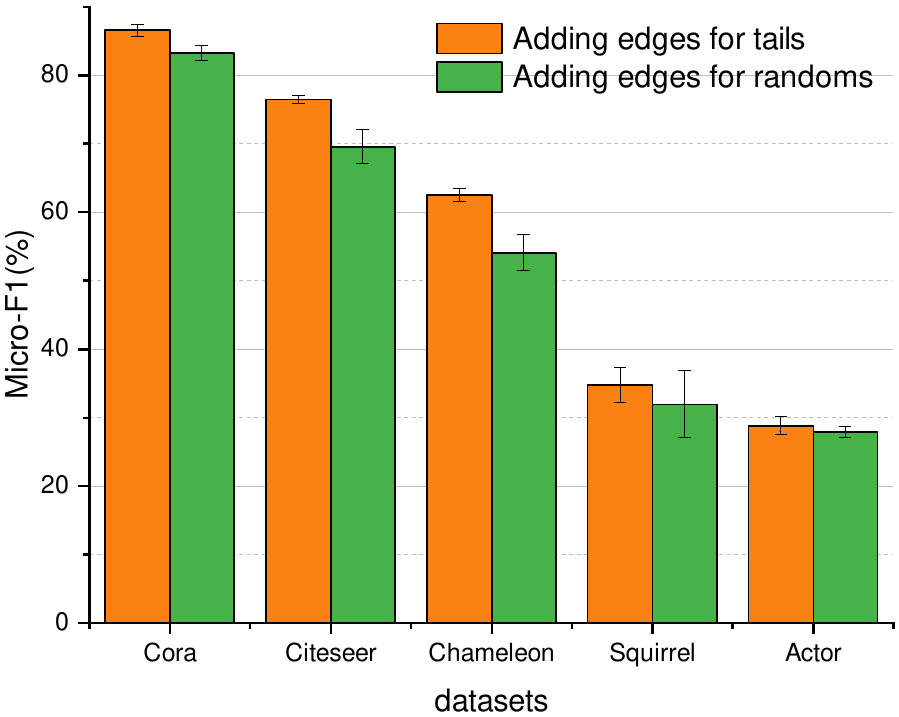}
		\label{adding_gat}
	}
	\subfigure[Denoising module in SAugGAT]
	{
		\includegraphics[width=0.45\linewidth]{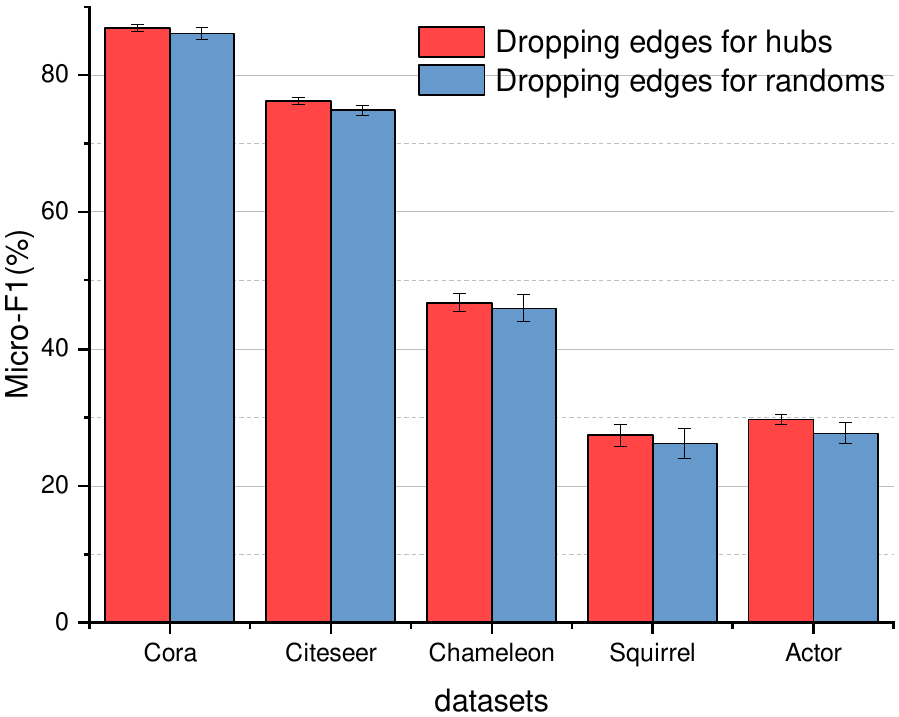}
		\label{denoise_gat}
	}
	\caption{Micro-F1 comparison of the model that adds/drops edges for tails/hubs and for random nodes}
	\label{ablation_nc}
\end{figure}

\begin{figure}[h]
	\centering
	\subfigure[Discovering module in SAugGCN]
	{
		\includegraphics[width=0.45\linewidth]{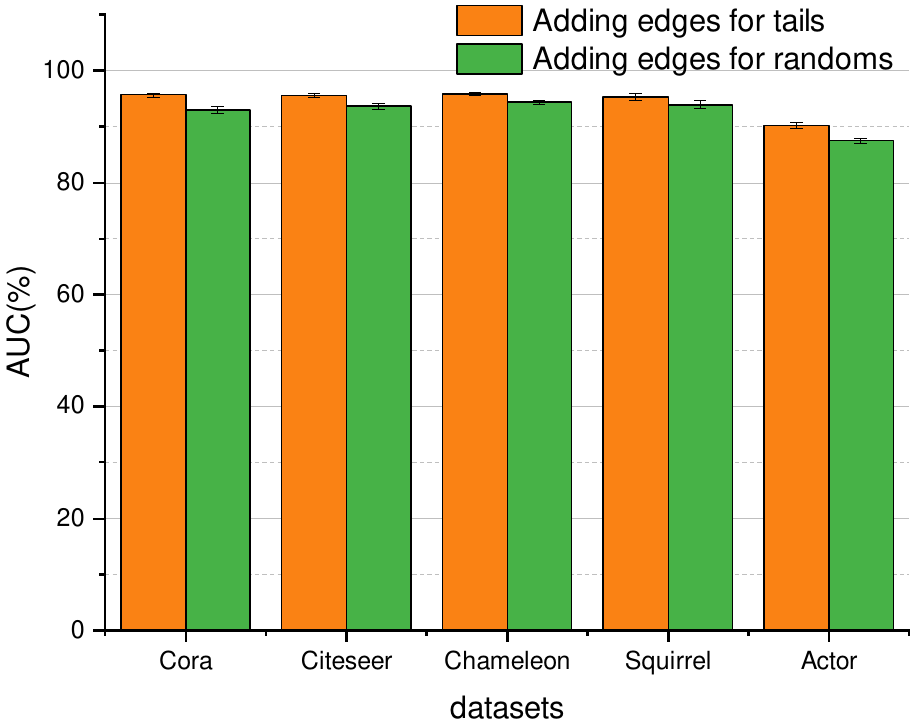}
		\label{lp_adding_gcn}
	}
	\subfigure[Denoising module in SAugGCN]
	{
		\includegraphics[width=0.45\linewidth]{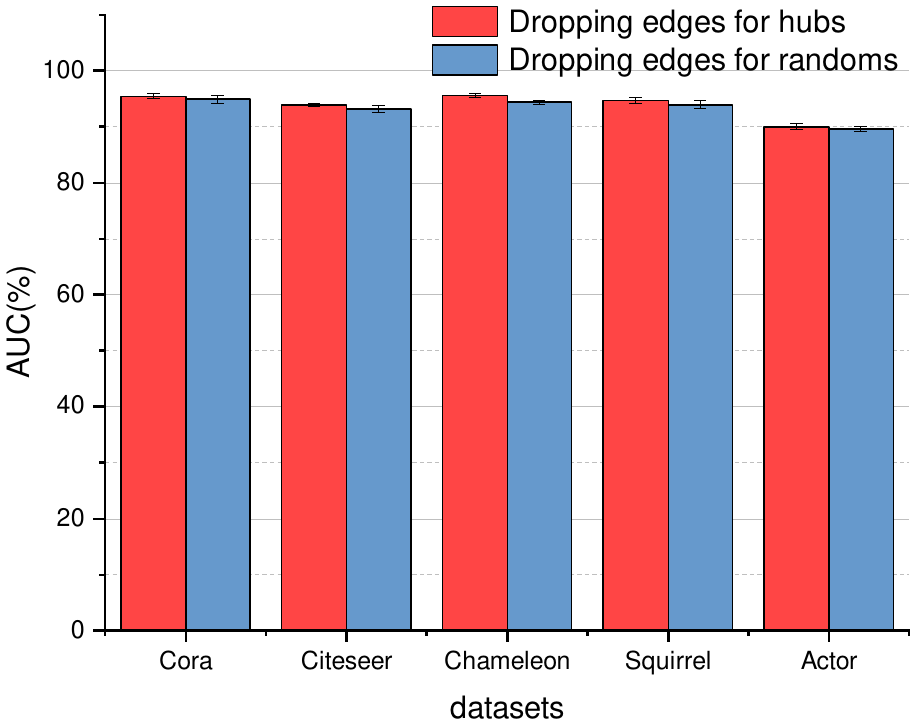}
		\label{lp_abaltion_gcn}
	}
	\\
	\subfigure[Discovering module in SAugSAGE]
	{
		\includegraphics[width=0.45\linewidth]{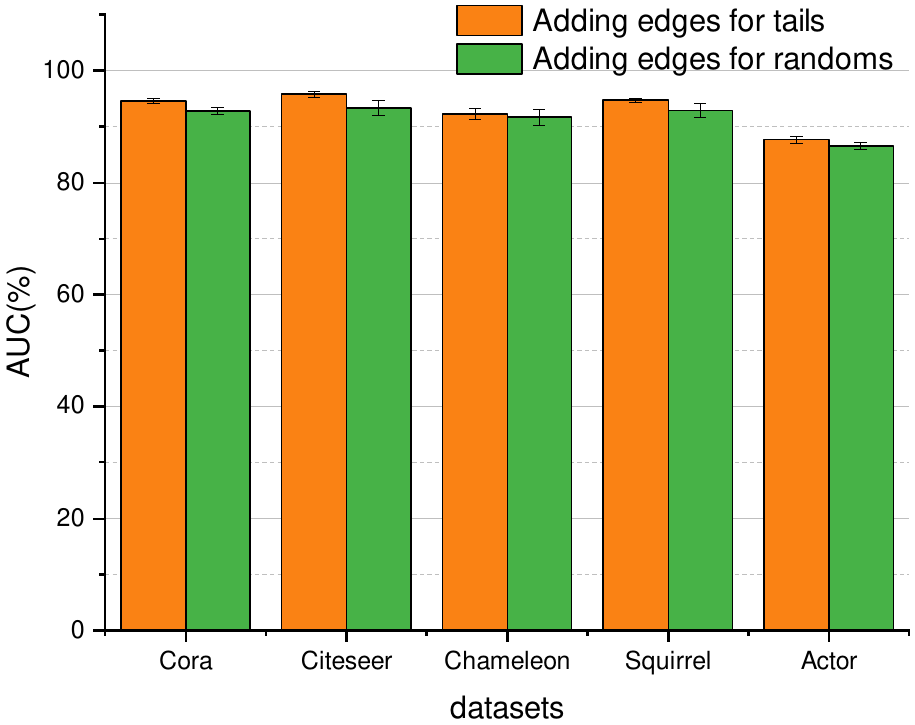}
		\label{lp_adding_sage}
	}
	\subfigure[Denoising module in SAugSAGE]
	{
		\includegraphics[width=0.45\linewidth]{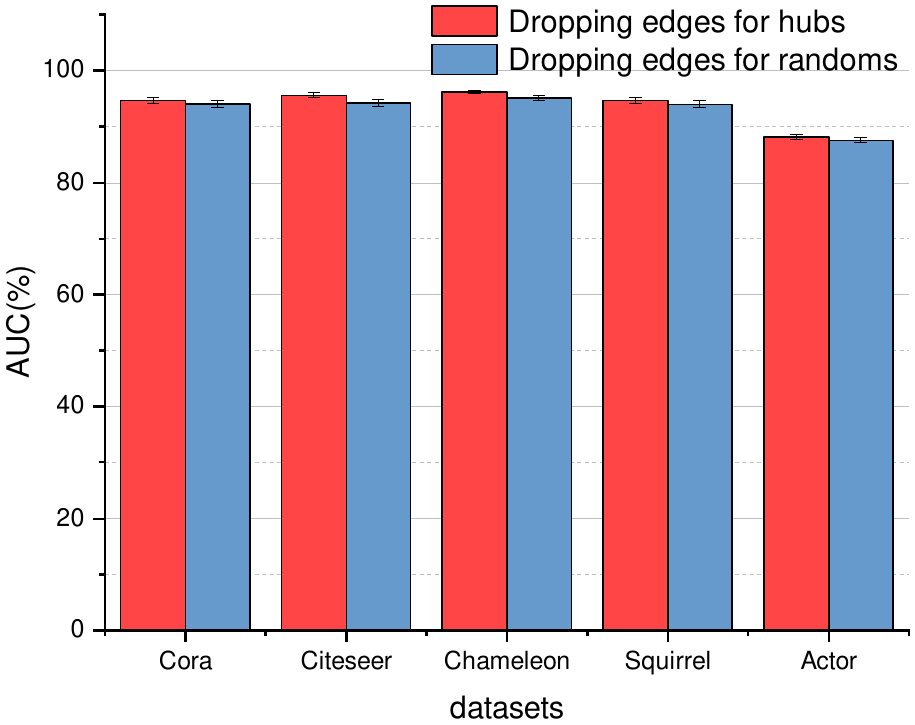}
		\label{lp_abalation_sage}
	}
	\\
	\subfigure[Discovering module in SAugGAT]
	{
		\includegraphics[width=0.45\linewidth]{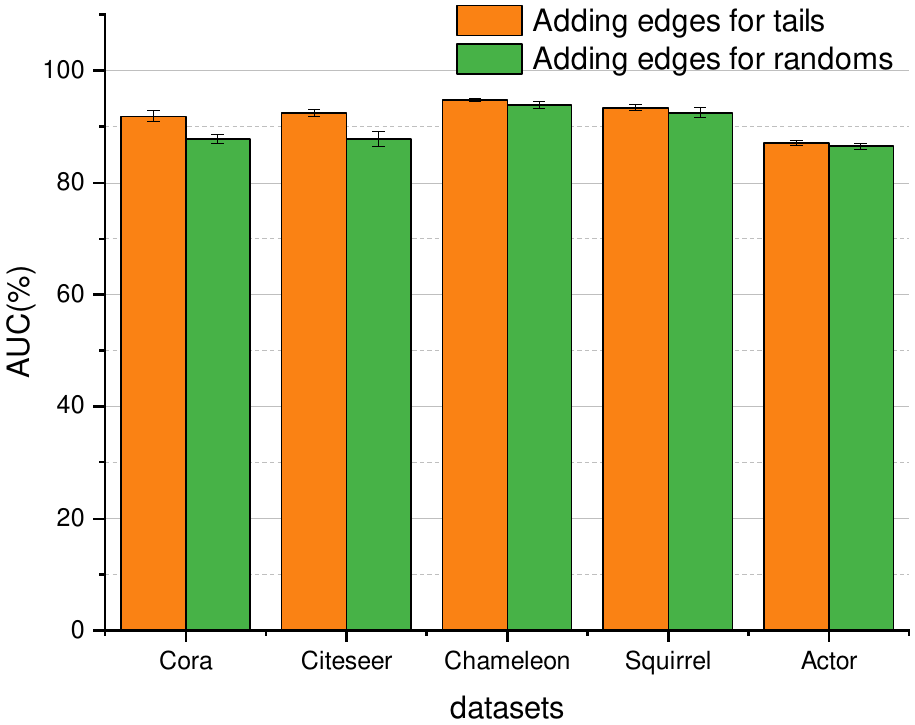}
		\label{lp_adding_gat}
	}
	\subfigure[Denoising module in SAugGAT]
	{
		\includegraphics[width=0.45\linewidth]{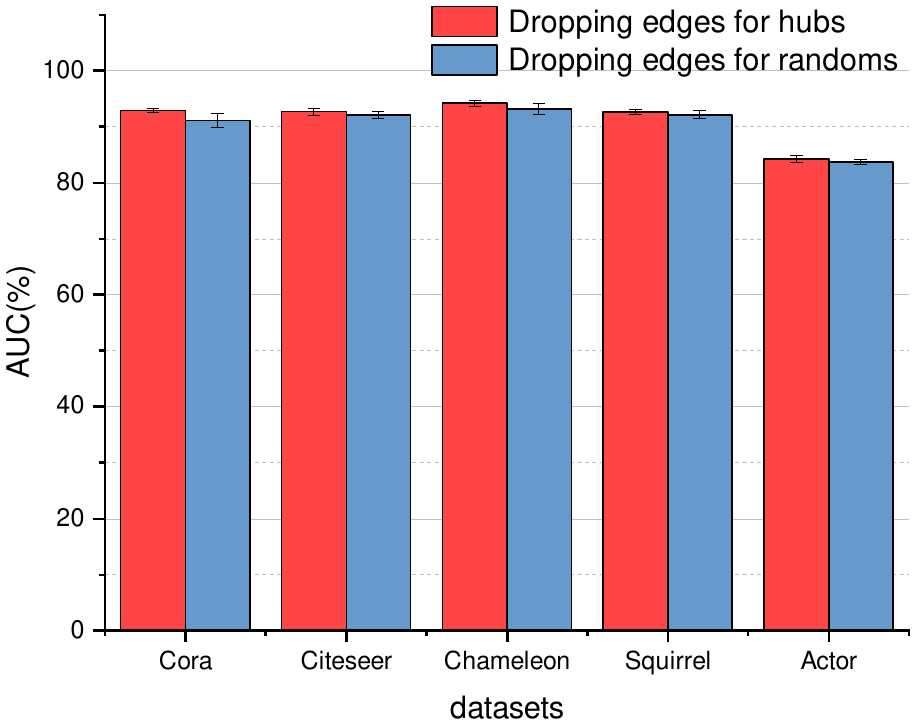}
		\label{lp_abaltion_gat}
	}
	\caption{AUC comparison of the model that adds/drops edges for tails/hubs and for random nodes on link prediction}
	\label{ablation_lp}
\end{figure}

\section{Complexity analysis}
Our SAug consists of four components: the Pagerank based hub-tail sampling strategy; the denosie module; the discovering module; the generative module. Given a graph $\mathcal{G}=(\mathcal{V},\mathcal{E}, \mathcal{A},\mathcal{X})$, the cost of the computating of Pagerank values is $O(\left|\mathcal{V}\right|)$ and the cost of sampling is $O(log(\left|\mathcal{V}\right|))$. In the denoise module and discovering module, a parallel computing using extra disc space is used to save the cost of simlarity computation $O(\left|\mathcal{V}\right|^2)$ to $O(\left|\mathcal{V}\right|)$. In the generative module, the cost of the generationr G~(MLP) is $O((l_{gen}-1)\left|\mathcal{V}_{gen}\right|H_{gen}^2+\left|\mathcal{V}_{gen}\right|H_{gen}d_\mathcal{X})=O(l_{gen}\left|\mathcal{V}_{gen}\right|H_{gen}d_\mathcal{X})$, where $l_{gen}$ is the numbers of layers of G, $H_{gen}$ is the hidden layer dimension size of G, $d_\mathcal{X}$ is the dimension size of feature $\mathcal{X}$, and $d_\mathcal{X}>H_{gen}$. The cost of the discriminator D~(GCN) is $O(l_{dis}\left|\mathcal{E}\right|H_{dis}+l_{dis}\left|\mathcal{V}\right|H_{dis}^2)$, where $l_{dis}$ is the numbers of layers of D, $H_{dis}$ is the hidden layer dimension size of D. The total cost of the generative module is $O(l_{gen}\left|\mathcal{V}_{gen}\right|H_{gen}d_\mathcal{X}+\lambda_2(l_{dis}\left|\mathcal{E}\right|H_{dis}+l_{dis}\left|\mathcal{V}\right|H_{dis}^2)=O(l_{gen}\left|\mathcal{V}_{gen}\right|H_{gen}d_\mathcal{X})$ where $\lambda_2$ is the number of training steps of D during the complete training of G. $\lambda_2$ is a constant number used to avoid the over-fitting of G and D since D~(GNNs) outperforms G~(MLP). Therefore, the cost of D can be removed for time complexity calculation. The total cost of SAug is $O(\left|\mathcal{V}\right|+log(\left|\mathcal{V}\right|)+\left|\mathcal{V}\right|+l_{gen}\left|\mathcal{V}_{gen}\right|H_{gen}d_\mathcal{X})=O(l_{gen}\left|\mathcal{V}_{gen}\right|H_{gen}d_\mathcal{X})$.

\section{Additional Experimental Results\label{additional_experiments}}
We provide some complete experimental results, including the complete results on tail node classification with GraphSAGE and GAT shown in Table~\ref{gnn_variants}, and the complete results on overall classification shown in Table~\ref{overall_nc}. Moreover, the complete ablation study results are shown in Figure~\ref{ablation_nc} and Figure~\ref{ablation_lp}.

\begin{table*}[htbp]
	\centering
	\resizebox{\textwidth}{!}
	{
		\begin{tabular}{l*{10}{c}}
			\toprule
			\multirow{2}*{Methods} &\multicolumn{2}{c}{Cora} &\multicolumn{2}{c}{Citeseer} &\multicolumn{2}{c}{Chameleon} &\multicolumn{2}{c}{Squirrel} &\multicolumn{2}{c}{Actor}
			\\
			\cmidrule(lr){2-3}\cmidrule(lr){4-5}\cmidrule(lr){6-7}\cmidrule(lr){8-9}\cmidrule(lr){10-11}
			&Macro-F1 &Micro-F1 &Macro-F1 &Micro-F1 &Macro-F1 &Micro-F1 &Macro-F1 &Micro-F1 &Macro-F1 &Micro-F1
			\\
			\midrule
			GraphSAGE &86.0±0.8 &87.0±0.8 &69.3±0.4 &74.2±0.5 &46.4±1.2 &47.0±1.0 &32.0±1.2 &32.9±1.6 &35.1±1.5 &42.4±1.0
			\\
			SAGE w/ $Gen$ &85.9±0.9 &86.9±0.7 &70.0±1.2 &74.6±0.6 &46.4±2.6 &47.1±2.6 &32.09±1.6 &34.2±1.6 &35.3±1.1 &42.5±1.2
			\\
			\midrule
			GAugSAGE &\underline{91.9±0.6} &\underline{92.6±0.5} &\underline{74.9±0.7} &\underline{79.0±0.6} &52.7±1.4 &53.1±1.3 &39.6±2.1 &40.4±1.9 &45.1±1.2 &53.9±1.6
			\\
			TailSAGE &91.4±0.5 &92.1±0.6 &74.1±0.6 &78.5±0.8 &\underline{55.3±1.7} &\underline{55.9±1.4} &\underline{42.9±2.0} &\underline{43.7±2.3} &\underline{51.7±1.9} &\underline{60.4±1.5}
			\\
			\midrule
			%Ours(Denoise) &86.1±0.7 &86.9±0.4 &70.7±0.6 &75.0±0.7 &46.7±1.2 &47.6±1.4 &32.1±0.9 &34.3±1.6 &35.4±1.1 &42.9±1.0
			SAug$_\mathrm{thr}$ w/o $Gen$ &\textbf{93.3±0.6} &\textbf{94.0±0.4} &\textbf{78.8±0.8} &\textbf{81.8±0.7} &60.7±2.0 &60.7±1.0 &44.3±0.8 &44.4±0.9 &\textbf{56.6±0.9} &64.4±0.6
			\\
			SAug$_{thr}$ &93.1±0.6 &93.8±0.5 &78.6±0.8 &81.6±0.7 &\textbf{61.1±1.2} &\textbf{61.3±1.5} &\textbf{44.7±1.6} &\textbf{44.9±1.7} &55.7±1.5 &\textbf{64.9±1.2}
			\\
			SAug$_{top}$ w/o $Gen$ &90.6±0.3 &91.3±0.3 &78.7±0.7 &80.3±0.6 &48.8±2.7 &49.9±2.5 &34.3±2.1 &35.2±2.3 &40.6±0.8 &48.1±0.8
			\\
			SAug$_{top}$ &90.8±0.5 &91.4±0.4 &78.6±0.7 &80.4±0.6 &48.6±1.5 &50.1±1.5 &33.9±1.5 &34.8±1.4 &40.7±0.7 &47.7±1.4
			\\
			\midrule
			\midrule
			GAT &84.7±1.1 &86.1±0.9 &71.7±1.5 &75.5±1.1 &44.7±2.0 &45.9±1.9 &28.7±2.1 &30.5±2.8 &20.6±1.1 &29.0±1.0
			\\
			GAT w/ $Gen$ &84.8±1.1 &86.0±0.9 &72.0±1.2 &75.9±0.9 &46.0±1.3 &47.4±1.7 &28.5±2.2 &30.2±2.9 &20.7±1.0 &29.4±0.9
			\\
			\midrule
			GAugGAT &\underline{85.6±0.5} &\underline{87.0±0.7} &72.9±0.4 &\underline{77.1±0.6} &51.4±0.6 &53.7±2.1 &29.5±2.4 &31.4±2.6 &21.2±1.6 &29.6±1.1
			\\
			TailGAT &84.9±0.4 &86.9±0.5 &\underline{73.0±0.7} &76.4±0.8 &\underline{53.9±2.1} &\underline{55.1±2.0} &\textbf{31.2±1.9} &\textbf{32.5±2.4} &\textbf{22.2±1.5} &\textbf{30.7±1.7}
			\\
			\midrule
			%Ours(Denoise) &85.3±1.0 &86.6±1.0 &71.6±1.6 &76.2±1.0 &45.5±1.5 &46.7±1.9 &25.5±2.1 &27.4±2.8 &20.4±1.0 &29.0±0.8
			SAug$_{thr}$ w/o $Gen$ &85.9±0.6 &87.2±0.6 &\textbf{74.3±0.5} &77.5±0.4 &\textbf{54.6±1.7} &\textbf{56.1±1.9} &\underline{30.7±2.1} &\underline{32.0±2.8} &\underline{21.7±1.3} &\underline{30.2±1.3}
			\\
			SAug$_{thr}$ &\textbf{86.1±0.9} &\textbf{87.4±0.7} &74.3±1.0 &\textbf{77.9±0.8} &53.6±2.1 &54.8±2.1 &29.5±2.8 &31.3±2.5 &21.0±1.0 &29.8±0.6
			\\
			SAug$_{top}$ w/o $Gen$ &85.9±0.7 &87.0±0.9 &72.9±1.0 &76.2±0.9 &52.6±2.0 &54.2±2.0 &24.9±3.1 &26.6±2.9 &20.6±0.7 &29.7±0.9
			\\
			SAug$_{top}$ &85.7±0.7 &86.8±0.8 &72.9±1.0 &76.2±0.9 &52.5±1.7 &54.2±2.0 &28.7±2.9 &30.5±2.8 &20.5±1.1 &29.5±0.9
			\\
			\bottomrule
		\end{tabular}
	}
	\caption{Micro-F1 scores on tail node classification with other GNN variants}
	\label{gnn_variants}
\end{table*}

\begin{table*}[htbp]
	\centering
	\resizebox{\textwidth}{!}
	{
		\begin{tabular}{l*{10}{r}}
			\toprule
			\multirow{2}*{Methods} &\multicolumn{2}{c}{Cora} &\multicolumn{2}{c}{Citeseer} &\multicolumn{2}{c}{Chameleon} &\multicolumn{2}{c}{Squirrel} &\multicolumn{2}{c}{Actor}
			\\
			\cmidrule(lr){2-3}\cmidrule(lr){4-5}\cmidrule(lr){6-7}\cmidrule(lr){8-9}\cmidrule(lr){10-11}
			&Macro-F1 &Micro-F1 &Macro-F1 &Micro-F1 &Macro-F1 &Micro-F1 &Macro-F1 &Micro-F1 &Macro-F1 &Micro-F1
			\\
			\midrule
			GCN &79.3±0.5 &80.1±0.6 &60.2±1.5 &63.1±0.9 &33.7±1.0 &35.4±1.1 &21.9±1.0 &23.1±1.2 &22.9±0.8 &23.9±1.0
			\\
			GCN w/ $Gen$ &80.2±0.7 &80.9±0.7 &60.7±1.1 &63.4±1.2 &34.4±1.3 &36.1±0.9 &22.4±1.4 &23.4±1.7 &23.4±1.5 &24.5±0.9
			\\
			\midrule
			GAug &82.1±0.4 &83.6±0.5 &\underline{66.4±0.8} &\underline{69.3±0.7} &32.9±1.4 &37.1±1.3 &22.4±1.5 &\underline{24.7±1.4} &23.1±1.1 &24.8±1.2
			\\
			CenGCN &\underline{82.7±0.6} &\underline{84.1±0.5} &64.4±0.9 &68.1±1.1 &\underline{36.4±1.4} &\underline{38.1±1.6} &\underline{23.2±1.2} &24.5±1.1 &\underline{25.4±0.9} &\underline{26.2±1.6}
			\\
			\midrule
			SAug$_{thr}$ w/o $Gen$ &85.2±0.6 &\textbf{86.4±0.5} &69.2±1.2 &71.9±1.3 &38.2±1.1 &39.1±1.4 &23.6±1.2 &\textbf{25.4±0.8} &\textbf{26.8±1.3} &28.4±0.9
			\\
			SAug$_{thr}$ &\textbf{85.6±0.8} &86.2±0.6 &\textbf{69.6±1.1} &\textbf{72.1±0.8} &38.6±1.3 &39.7±1.4 &\textbf{23.7±1.1} &24.1±1.0 &26.5±1.1 &\textbf{28.7±1.2}
			\\
			SAug$_{top}$ w/o $Gen$ &84.1±0.5 &85.2±0.4 &68.5±0.9 &70.4±0.9 &38.9±0.9 &40.0±1.7 &22.4±0.9 &24.3±1.1 &26.4±0.7 &27.9±1.0
			\\
			SAug$_{top}$ &84.7±0.7 &85.4±0.8 &68.9±1.3 &70.9±1.4 &\textbf{39.2±1.2} &\textbf{40.5±1.8} &22.7±1.4 &24.7±1.2 &26.5±1.4 &28.1±1.4
			\\
			\midrule
			\midrule
			GraphSAGE &79.7±1.4 &81.9±1.3 &68.5±0.6 &72.5±0.9 &33.5±0.8 &48.0±1.6 &24.6±0.5 &34.0±0.9 &26.8±1.3 &30.9±1.1
			\\
			GraphSAGE w/ $Gen$ &80.4±1.1 &82.1±1.0 &68.9±0.9 &72.6±1.2 &33.7±1.4 &48.9±1.2 &24.9±1.1 &34.4±1.2 &27.3±1.7 &31.2±1.6
			\\
			GAug &\underline{80.9±0.7} &\underline{83.2±0.4} &\underline{71.4±0.5} &\underline{75.7±0.7} &\underline{38.4±1.6} &\underline{52.0±1.2} &\underline{28.1±0.9} &\underline{37.1±1.1} &\underline{29.5±1.4} &\underline{33.4±1.2}
			\\
			\midrule
			SAug$_{thr}$ w/o $Gen$ &90.5±0.9 &91.2±1.3 &76.1±0.6 &78.7±0.9 &55.1±1.4 &55.8±0.9 &\textbf{46.2±1.1} &46.4±1.1 &29.1±1.2 &35.7±1.1
			\\
			SAug$_{thr}$ &\textbf{90.8±1.0} &\textbf{91.6±1.2} &\textbf{76.3±0.5} &\textbf{79.1±0.6} &55.4±1.1 &56.1±1.2 &46.0±1.6 &\textbf{46.7±0.8} &\textbf{29.5±0.8} &\textbf{36.1±0.7}
			\\
			SAug$_{top}$ w/o $Gen$ &89.7±1.4 &90.4±1.4 &75.6±0.7 &78.4±0.5 &\textbf{55.8±1.2} &56.1±1.4 &45.7±1.4 &45.4±1.3 &29.0±1.3 &35.1±0.9
			\\
			SAug$_{top}$ &90.1±0.9 &90.7±1.1 &75.9±0.8 &78.1±0.7 &55.4±1.3 &\textbf{56.3±1.0} &45.9±1.5 &45.9±1.2 &28.9±1.4 &35.8±1.2
			\\
			\midrule
			\midrule
			GAT &75.5±1.4 &76.4±1.6 &61.0±1.1 &64.4±1.5 &46.3±1.6 &47.5±1.3 &30.6±2.3 &31.4±1.7 &19.1±1.0 &26.9±1.0
			\\
			GAT w/ $Gen$ &75.9±1.7 &76.5±1.4 &61.4±1.7 &64.7±1.1 &46.7±1.1 &47.7±1.6 &30.9±2.1 &31.6±1.9 &19.5±1.6 &27.1±1.4
			\\
			GAug &76.5±1.2 &\underline{77.3±0.9} &62.9±1.5 &\underline{66.7±1.2} &47.9±1.4 &\underline{49.7±1.6} &31.9±2.4 &\underline{32.4±1.5} &19.9±1.1 &\underline{27.4±1.2}
			\\
			\midrule
			SAug$_{thr}$ w/o $Gen$ &76.7±1.1 &\textbf{77.7±1.0} &64.2±1.6 &67.4±1.3 &55.6±1.8 &56.3±1.7 &75.0±2.0 &33.8±1.2 &20.0±1.7 &27.4±1.5
			\\
			SAug$_{thr}$ &\textbf{76.9±1.7} &77.5±1.3 &\textbf{64.5±1.6} &\textbf{67.7±1.8} &\textbf{56.0±1.7} &\textbf{56.6±1.5} &75.4±1.9 &33.4±1.2 &\textbf{20.2±1.2} &\textbf{27.6±1.1}
			\\
			SAug$_{top}$ w/o $Gen$ &75.7±1.2 &76.4±1.4 &63.5±1.2 &66.9±1.4 &55.4±1.2 &54.8±1.9 &75.1±1.7 &\textbf{33.9±1.1} &19.4±1.3 &27.1±1.2
			\\
			SAug$_{top}$ &75.4±1.4 &76.9±1.0 &63.8±1.7 &67.1±1.8 &55.6±1.9 &55.1±1.4 &\textbf{75.5±1.4} &33.7±1.7 &19.7±1.6 &27.3±1.4
			\\
			\bottomrule
		\end{tabular}
	}
	\caption{Micro-F1 scores on overall node classification. In each comparison, the best result is bolded and the second bestresult is underlined.}
	\label{overall_nc}
\end{table*}

%% The file named.bst is a bibliography style file for BibTeX 0.99c
\bibliographystyle{named}
\bibliography{ijcai23}

\end{document}